\documentclass[10pt,twocolumn,letterpaper]{article}

\usepackage{multirow}
\usepackage{adjustbox}
\usepackage[font=small]{subcaption}

\usepackage{cvpr}
\usepackage{times}
\usepackage{epsfig}
\usepackage{graphicx}
\usepackage{amsmath}
\usepackage{amssymb}

\usepackage[pagebackref=true,breaklinks=true,letterpaper=true,colorlinks,bookmarks=false]{hyperref}

\cvprfinalcopy 

\def\cvprPaperID{715} 

\ifcvprfinal\fi

\begin{document}

\title{A New Convolutional Network-in-Network Structure and Its Applications in Skin Detection, Semantic Segmentation, and Artifact Reduction}

\author{Yoonsik Kim, Insung Hwang, Nam Ik Cho\\
Dept. of Electrical and Computer Engineering\\
Seoul National University, Seoul, Korea\\
{\tt\small terryoo@ispl.snu.ac.kr, coee55@gmail.com, nicho@snu.ac.kr}
}

\maketitle

\begin{abstract}
The inception network has been shown to provide good performance on image 
classification problems, but there are not much evidences that it is also effective 
for the image restoration or pixel-wise labeling problems. 
For image restoration problems, the pooling is generally not used because 
the decimated features are not helpful for the reconstruction of an image as the output.
Moreover, most deep learning architectures for the restoration problems do not use dense prediction 
that need lots of training parameters.
From these observations, for enjoying the performance of inception-like structure on 
the image based problems we propose a new convolutional network-in-network 
structure. The proposed network can be considered a modification of inception 
structure where pool projection and pooling layer are removed for maintaining 
the entire feature map size, and a larger kernel filter is added instead.
Proposed network greatly reduces the number of parameters on account of removed
dense prediction and pooling, which is an advantage, but may also reduce the receptive field in each layer. 
Hence, we add a larger kernel than the original inception structure for not increasing 
the depth of layers. The proposed structure is applied to typical image-to-image 
learning problems, i.e., the problems where the size of input and output are same 
such as skin detection, semantic segmentation, and compression artifacts reduction. 
Extensive experiments show that the proposed network brings comparable or 
better results than the state-of-the-art convolutional neural networks for 
these problems.	
\end{abstract}

\section{Introduction}
In recent year, various convolutional networks have been developed for 
the applications from low to high level vision problems such as 
classification, object detection, image restoration and segmentation problems~\cite{krizhevsky2012imagenet,simonyan2014very,
szegedy2015going,7115171,wang2015deep,girshick2014rich,long2015fully,wang2015designing}.
Among many problems, this paper focuses on developing a 
convolutional network that works for image segmentation and filtering problems.
For the semantic segmentation, Long \etal proposed a 
convolutional network~\cite{long2015fully}, which adopts a classification network 
as an encoder part for exploiting pre-trained features. Then the entire 
image is observed by using a pooling and converting the fully connected 
layer to a convolution layer. However, the labeling results seem somewhat 
coarse due to the limitation of reconstruction part which is consist of 
deconvolution layer and up sampling layer. Some researchers tried to alleviate 
this weakness by cascade deconvolutional network training~\cite{noh2015learning} 
or indexing up sampling layer~\cite{badrinarayanan2015segnet}. 

\begin{figure}[t]
	\centering		
		\includegraphics[width=0.95\linewidth]{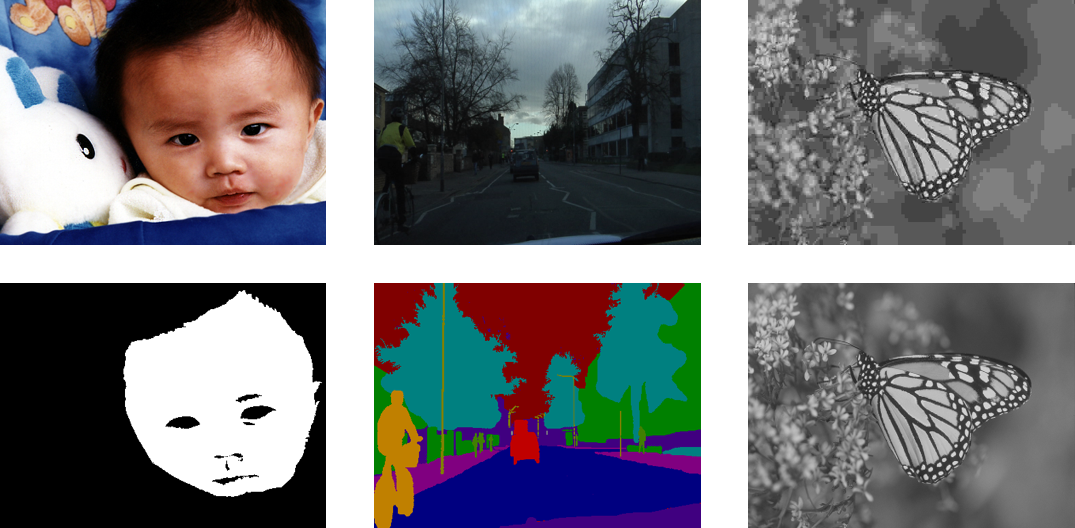}			
	\caption{Example of image-to-image deep learning problems where dimension of input and output (label) is same: (from left to right) skin detection, semantic segmentation, and compression artifact reduction}
	\label{fig:Skin Inception23123}		
\end{figure}

Meanwhile, convolutional network approaches for lower class pixel-wise 
classification such as salient region detection~\cite{wang2015deep,zhao2015saliency}, 
surface normal classification, and edge orientation classification \cite{wang2015designing} 
have also been developed, which may be regarded as specific applications of 
semantic segmentation. In these works, they tried to find appropriate convolutional networks 
and post processing steps to the convolutional network output. 
In many of the above stated works and many other recent works,
a classification network model~\cite{liu2016dhsnet} is usually adopted
for exploiting the pre-trained features. 
However, it is not clear whether adopting a classification network
is also effective for the pixel-wise labeling problems with small number of labels.
Further, for the problems where the input and output have the
same dimension as in many of image restoration and labeling problems discussed so far,
since successive dimension reduction by pooling reduces specific pixel information,
it is not helpful at the image reconstruction stage.

From these reasoning, we propose a new convolutional network-in-network structure
that can be applied the image-to-image deep learning problems as in 
Figure~\ref{fig:Skin Inception23123}.
The proposed structure can be considered a modification of inception network, 
where the pooling is
removed and a larger kernel is added instead.
Our main contributions are summarized as follows.

\begin{enumerate}
\item [$\bullet$]We propose a new inception-like convolutional network-in-network structure, which 
consists of convolution and rectified linear unit (ReLU) layers only.  That is, we exclude 
pooling and subsampling layer that reduce feature map size, because decimated features 
are not helpful at the reconstruction stage. Hence, it is able to do one-to-one (pixel wise) 
matching at the inner network and also intuitive analysis of feature map correlation.	
	
\item [$\bullet$] Proposed architecture is applied to several pixel-wise labeling and restoration
problems and it is shown to provide comparable or better performances compared to 
the state-of-the-art methods. 
\end{enumerate}

In the rest of this paper, overview of related methods is introduced in section 2, 
and the proposed structure is presented in section 3. 
Analysis and comparison are provided in section 4 and this paper is concluded in section 5. 

\begin{figure*}[ht!]
\includegraphics[width=\textwidth]{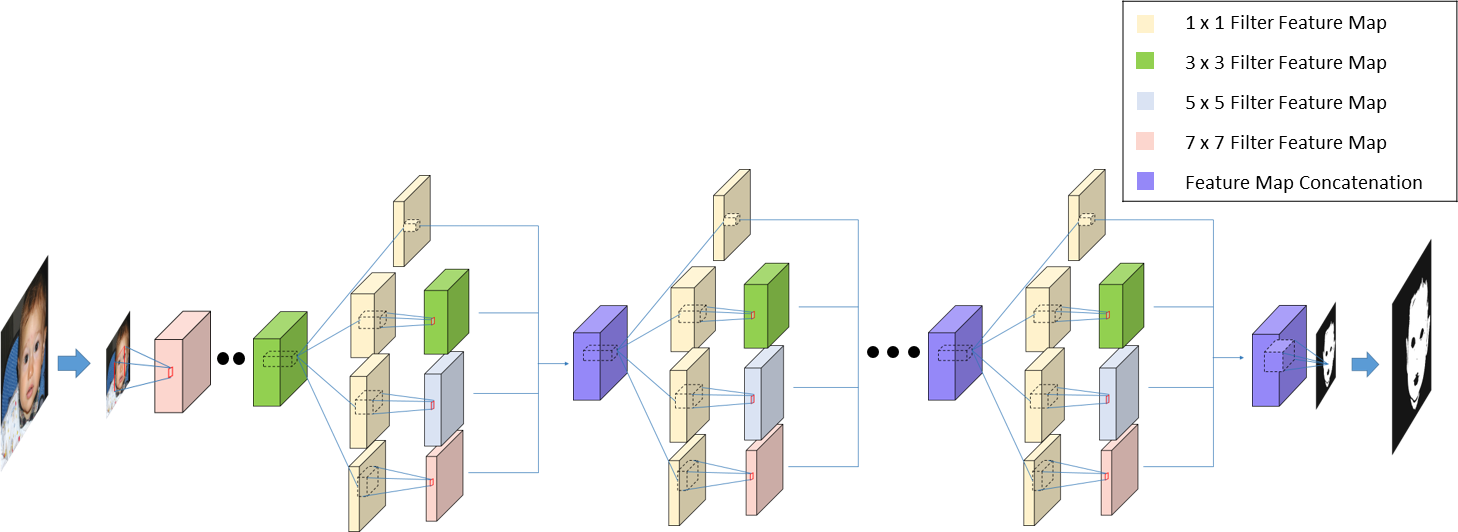}
\caption{Proposed Network-in-Network Structure applied to skin detection. The architecture is 
composed of 4 convolution layers with nonlinear layers and 8 modified inception modules. 
Input image goes through the convolutional network to be a skin probability map at the output.}
\label{fig:Overview Global Network}
\end{figure*}

\section{Related Works}
Since the proposed architecture is very simple in that it uses only convolution layer and
ReLU, which are widely known and used, we skip the review of these. Instead we 
review the areas that we apply our architecture: skin detection, semantic segmentation, and compression artifacts reduction as illustrated in Figure~\ref{fig:Skin Inception23123}.

\subsection{Skin Detection}
Skin detection is to locate skin pixels or regions in images, which is an important  
pre-processing step in many image processing and computer vision tasks.
For example, it can be used in image enhancement~\cite{zafarifar2013application}, 
face and human detection~\cite{hsu2002face}, gesture analysis~\cite{Francke2007real}, 
pornographic contents filtering~\cite{jones2002statistical}, 
surveillance systems~\cite{zhang2009head}, etc. 
Hence there have been a large number of skin detection algorithms, 
which is well summarized in~\cite{kakumanu2007survey,vezhnevets2003survey,bianco2015adaptive}.
According to these works, the conventional methods find skin pixels that fit to parametric models  ~\cite{hsu2002face,yang1998gaussian,lee2002elliptical}, 
nonparametric models~\cite{sigal2004skin,jones2002statistical,khan2010skin}, 
or some skin-cluster defined regions in certain color spaces~\cite{yang1998skin}.
There are also some methods that detect human related features (hands, faces, and body) 
first for finding skin pixels~\cite{liao2008estimation, bianco2015adaptive},
while most of the existing works detect skin pixels for finding human. 
In addition, there are neural network methods that try to find skin areas 
such as adaptive neural networks~\cite{phung2001universal} and self-organizing 
maps~\cite{BMVC.15.51}.
But they are quite old dated and
do not use open database so that direct comparison with existing neural network methods seems
very difficult. Even though there have been many algorithms,
the skin detection is still considered a challenging problem due to diverse 
variations of images, illumination conditions, skin color variations 
of races and makeup, and skin-like background.

\subsection{Semantic Segmentation}
Pixel-wise semantic segmentation problem is to classify each pixel, \ie, to 
label a pixel as one of classes. 
As some examples of non-convolutional network approaches,
hand-crafted features are extracted from image patches and they are
trained by random forest classifier
~\cite{shotton2008semantic,brostow2008segmentation} or boosting~\cite{sturgess2009combining}. 
After the success of convolutional network to classification problems~\cite{krizhevsky2012imagenet,simonyan2014very,szegedy2015going}, 
many researchers tried to apply convolutional network to semantic segmentation 
problem with patch based method~\cite{NIPS2012_4741,farabet2013learning,icml2014c1_pinheiro14}.
More recently, FCN~\cite{long2015fully,7478072} which contains classification network 
and deconvolution layers as encoder and decoder has been proposed. The encoder is a modified VGG~\cite{simonyan2014very} network which converts the original fully connected layer to a 
convolution layer. Therefore, the encoder fully reflects entire image and thus obtained the best 
performance on benchmark PASCAL VOC 2012 dataset. 
Many researchers also proposed new networks to improve the performance of FCN 
~\cite{noh2015learning,badrinarayanan2015segnet}.

\subsection{Reduction of Compression Artifacts}

Lossy compression leads to image degradation which is called compression artifacts 
such as blocking, ringing, and blurring artifacts. 
Conventional non-convolutional network approaches manipulate the artifacts in spatial domain~\cite{reeve1984reduction,Xu:2011:ISV:2070781.2024208,wang2013adaptive} or 
frequency domain~\cite{foi2007pointwise}, and more recent work using convolutional network
has shown that deep learning approach provides better results than the
above state methods~\cite{dong2015compression}.
After this work, many researchers also proposed
various kinds of artifact reduction networks~\cite{svoboda2016compression,yu2016deep}.
 
 \section{Proposed Convolutional Network}
After the convolutional network with inception module~\cite{szegedy2015going} has 
been developed, there have been many variants and applications adopting
the inception module for the classification problems. However, 
it seems there are not much evidences that the inception based network
is also effective for the image-to-image problems. To be precise,
although the inception module has the advantage that it extracts scale
invariant features by using multi-size kernels and thus shows
good performance on classification problems, it cannot be effectively
used for image-to-image problems due to decimated features by
pooling layers. As an example with the image restoration problems,
the pooling and decimation are generally not used because
we need to keep the features from the original size image for the
successful reconstruction of output image~\cite{Kim_2016_CVPR,dong2015compression,7115171}.
Further the restoration problems do not employ dense prediction 
method which is usually used in classification problems, because dense 
prediction needs lots of train parameters.
Hence, for taking advantage of network-in-network
structure for the image-to-image problems,
we propose a  new convolutional module which can be
considered a modification of inception module. 
Specifically, we remove pooling in the inception module and add a larger size kernel instead
to widen the receptive field which might have been reduced by the
removal of pooling.

The overall convolutional network architecture using the proposed module is shown in
in Figure~\ref{fig:Overview Global Network}, which is consisted of
several convolutional modules and the cascade of proposed multi-kernel
network-in-network modules. The figure shows the application of the
architecture in skin detection problem, and the same architecture is
also applied to semantic segmentation and compression artifacts removal 
except that the input can be a patch or decimated image, and also that
the depth of the network can be changed. In the following, we explain
the details of proposed module and overall architecture.

\subsection{Proposed Inception Module}

Christian \etal~\cite{szegedy2015going} proposed GoogLeNet with an inception module 
that extracts multi-scale feature maps by using multiple filters with different sizes, 
\eg, $1\times1$, $3\times3$, $5\times5$, and also by using pool projection layer 
which is composed of $1\times1$ convolution filter and max pooling to make pooling 
feature maps. To reduce the number of parameters, an additional $1\times1$ 
convolution filter (named as \textquotedblleft reduction layer \textquotedblright) is followed by comparably expensive 
$3\times3$ and $5\times5$ filters. All the feature maps extracted from the filters are 
integrated at the end of the inception module.
Since object regions change considerably in sizes according to scales and perspective views,
the inception module is designed to reflect diverse scales of objects by incorporating
multi-size filters. A filter with small kernel has the role of detecting small object 
regions while a larger filter contributes to not only detecting large object regions 
but also effectively suppressing false positive regions that have similar properties 
as the targeting object.

\begin{figure}[t]
	\centering
	\begin{subfigure}[]{0.5\linewidth}
		\centering
		\includegraphics[width=0.95\linewidth]{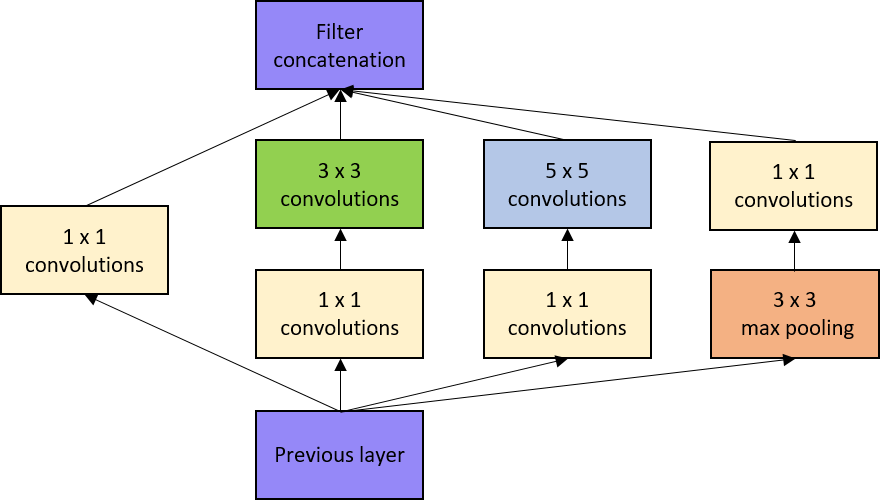}
		\caption{\small Original Inception}
	\end{subfigure}%
	\begin{subfigure}[]{0.5\linewidth}
		\centering
		\includegraphics[width=0.95\linewidth]{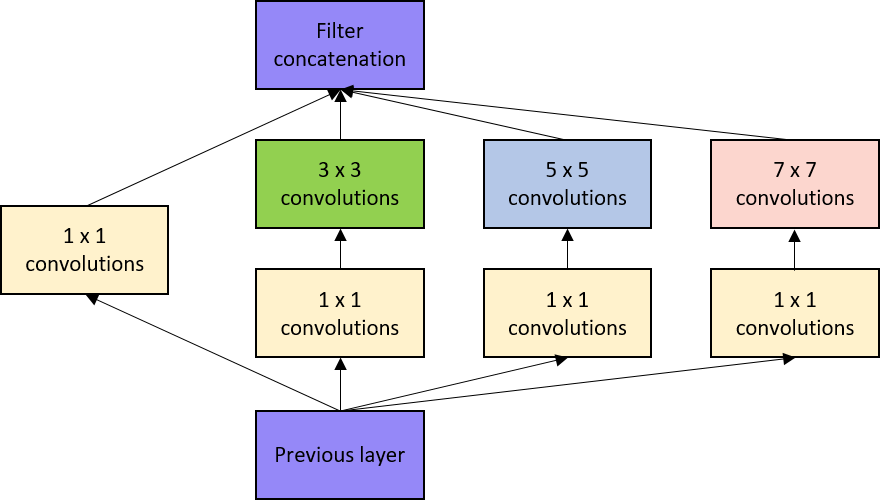}
		\caption{\small Modified Inception}		
	\end{subfigure}	
	\vspace{0.3cm}
	\caption{Inception modules}
	\label{fig:Skin Inception}		
\end{figure}

\begin{figure}[t]
\includegraphics[width=\linewidth]{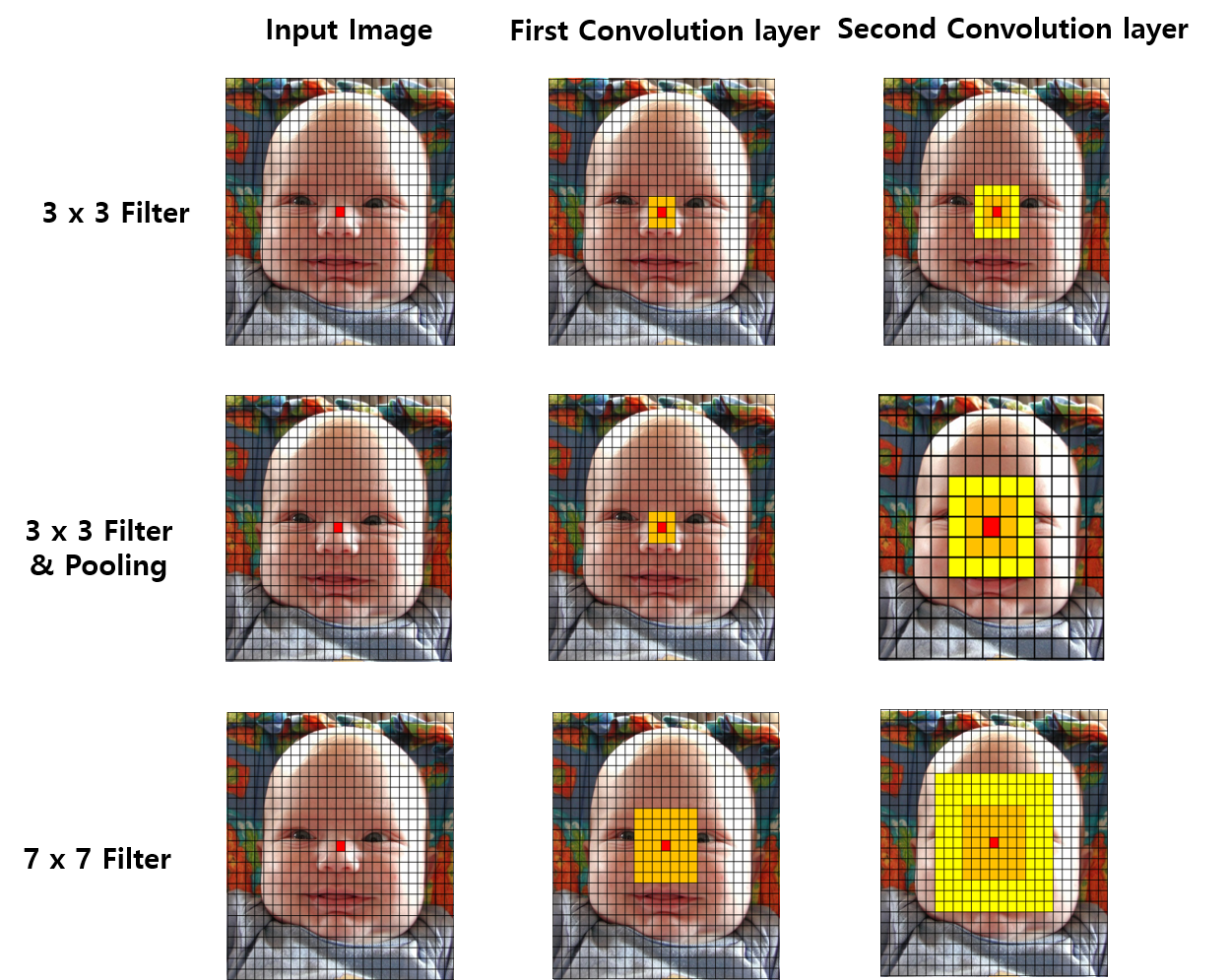}
\caption{Example of the range of input pixels according to the filter size and pooling layer. 
It compares the receptive field when the network consists of 
$3\times3$ kernel (first row), $3\times3$ kernel with pooling (second row) 
and $7\times7$ kernel (third row). 
First column is the input,  second is the receptive area at the first layer,
and the third shows the receptive field after the second layer with the
same size kernels or pooling.}
	\label{fig:Filter}
\end{figure}

We modify the inception module for image-to-image tasks by adding a larger kernel 
($7\times7$) and excluding the pool projection layer as shown in 
Figure~\ref{fig:Skin Inception} (b), where the original inception module is shown 
in Figure~\ref{fig:Skin Inception} (a) for comparison.
The original inception module~\cite{szegedy2015going} collects information from a wide range 
of images due to the overall pooling layers in the network, \ie, the pooling layers enlarge 
receptive fields by reducing the size of feature dimension.
However, since we remove the combination of pooling and
fully connected layers in order not to decimate the feature map,
we need to include a larger filter for keeping the receptive field large.
As a specific example for this, 
Figure~\ref{fig:Filter} shows how many pixels are involved in detecting 
a certain pixel according to filter size and the existence of the pooling layer.
The first row of Figure~\ref{fig:Filter} shows that the receptive range
becomes $5\times5$ pixels at the second layer when we use $3\times3$ kernels
consecutively. The second row shows that the receptive range becomes wider to
$10\times10$ (at the image before decimation) when we add pooling to the
above result. However, since we do not have pooling, we add a larger kernel ($7\times7$) 
to keep the receptive range as shown in the last row of Figure~\ref{fig:Filter}.

In summary, the proposed module is composed of a set of filters with different 
sizes ($1\times1$, $3\times3$, $5\times5$, $7\times7$) as illustrated in 
Figure~\ref{fig:Skin Inception}, each of which consists of 8, 32, 16, and 
8 filters respectively. We determine the number of filters in modified inception depending 
on the size of the filter in order to reduce the number of parameters.

\subsection{Overall Architecture}
The proposed method is applied to end-to-end mapping problems, 
which directly generate the overall output map.
The network is composed of deep 
convolution layers and proposed inception modules in order to exploit global 
evidence from the entire image. The number of overall parameters is 300K, 
which is much smaller than the same-depth inception networks adopted 
for classification problems.  

\begin{table*}[t]
	\centering
	\caption{Proposed Network-in-Network Architecture based on Modified Inception}
	
	\begin{adjustbox}{width=0.8\textwidth}
		\renewcommand{\arraystretch}{1.5}%
		\begin{tabular}{|c| c| c| c| c| c| c| c| c| c| c| c|}
			\hline
			type & \begin{tabular}{@{}c@{}}Kernel Size\end{tabular} & Output size & Depth & \# $1\times1$ & \begin{tabular}{@{}c@{}}\# $3\times3$ \\ Reduce \end{tabular}   & \# $3\times3$   & \begin{tabular}{@{}c@{}}\# $5\times5$ 5\\ Reduce \end{tabular} & \# $5\times5$  & \begin{tabular}{@{}c@{}}\# $7\times7$ \\ Reduce \end{tabular} & \# $7\times7$  & Params   \\ [0.5ex] 
			\hline\hline
			Convolution 1 &  $7\times7$  & H $\times$ W $\times$ 64 & 1 & & & & & & & &9K  \\ 
			\hline
			Convolution 2 &  $3\times3$  & H $\times$ W $\times$ 64 & 2 & &64 &64 & & & & &41K  \\		
			\hline
			Inception 1 &  & H $\times$ W $\times$ 64 & 2 & 8 &  32 & 32 & 16 & 16 & 8 & 8&23K  \\
			\hline
			Inception 2 &  & H $\times$ W $\times$ 64 & 2 & 8 &  32 & 32 & 16 & 16 & 8 & 8&23K \\
			\hline		
			Inception 3 &  & H $\times$ W $\times$ 64 & 2 & 8 &  32 & 32 & 16 & 16 & 8 & 8&23K \\
			\hline
			Inception 4&  & H $\times$ W $\times$ 64 & 2 & 8 &  32 & 32 & 16 & 16 & 8 & 8&23K  \\
			\hline
			Inception 5 &  & H $\times$ W $\times$ 64 & 2 & 8 &  32 & 32 & 16 & 16 & 8 & 8&23K  \\
			\hline
			Inception 6&  & H $\times$ W $\times$ 64 & 2 & 8 &  32 & 32 & 16 & 16 & 8 & 8&23K \\
			\hline
			Inception 7&  & H $\times$ W $\times$ 64 & 2 & 8 &  32 & 32 & 16 & 16 & 8 & 8&23K \\
			\hline
			Inception 8&  & H $\times$ W $\times$ 64 & 2 & 8 &  32 & 32 & 16 & 16 & 8 & 8&23K \\
			\hline
			Convolution 3&  $5\times5$  & H $\times$ W $\times$ 1 & 1 & & & & & & & &16K \\
			\hline
			Euclidean &  & H $\times$ W $\times$ 1 &  1 & & & & & & & & \\
			\hline
		\end{tabular}
	\end{adjustbox}	
	\label{table:kysymys}	
\end{table*}

\subsubsection{Input and Output Design for the Proposed Architecture}
As stated previously, we apply the proposed architecture to three problems: skin detection,
semantic segmentation and compression artifacts reduction. 
In the case of semantic segmentation, since the images in the dataset have
the same size, we just put the overall image as the input. In the
case of compression artifacts reduction, we set the input
and output as $37 \times 37$ image patches. For the
skin detection problem, the overall image should also be the
input because we wish to capture the wide range features rather
than just color and texture of small patches. However, since
the images in the dataset have various sizes and aspect ratios,
we design the input and output as follows:

\noindent \textbf{Input for the training}: Decimate the image to a fixed size such that smaller 
 side (horizontal/vertical) is reduced to 50, and stride the $50 \times 50$ image
 to other (larger side) direction. 
 
\noindent \textbf{Label image for the training}: Decimated and stridden
ground truth binary map in the same way as the input.

\noindent \textbf{Input for inference}: Image that is decimated such that smaller side (horizontal/vertical) is reduced to 50.

\noindent \textbf{Final output}: Probability or binary maps are obtained from the network,
and they are interpolated to the original size.

\subsubsection{Architecture Details for the Skin Detection Problem} 
We explain the details only for the skin detection problem, because other
problems use almost the same architecture.
The proposed network for the skin detection consists of 4 convolution layers 
and 8 modified inception modules where 
each module is composed of 2 convolution layers, so that the depth of the network 
adds up to 20 layers as described in Table~\ref{table:kysymys} where \textit{H}
 and \textit{W} are decimated height and width.
It is notated that ``$3\times3$ \textit{reduction}, $5\times5$ \textit{reduction}, 
and $7\times7$ \textit{reduction}'' are the reduction layers for $3\times3$, $5\times5$
 and $7\times7$ respectively.
The ReLU layer is used as an activation function 
which is defined as:
\begin{equation}
f(x) = max(0,x)
\end{equation}
which follows each convolution layer except for the last.

The proposed network can be divided into four parts. 
The first part consists of 3 convolution layers of $7\times7$, $1\times1$, and $3\times3$ kernels.
The second part is a set of modified inception modules which are described in section 3.1 where all  
modules are identical. The number of modified inception modules is set to 8 which will be 
discussed in section 4.2. Then, one convolution layer is used to integrate all the 
feature maps in a single 
channel which is a skin map produced by the proposed method.
At last, the loss function is included for the training phase.

\subsubsection{Loss function}
Given training images and ground truth $\left\{\mathbf{X}_i,\mathbf{Y}_i\right\}_{i=1}^{N}$, 
our goal is to learn the parameters that minimize the loss function which is defined 
as Euclidean loss:
\begin{equation}
\mathbf{L}(\Theta) = \frac{1}{N}\sum_{i=1}^{N} \| \mathbf{Y}_i - F(\mathbf{X}_i;\Theta) \|^2,
\end{equation}     
where $N$ is the number of training images and $\Theta$ is the parameters of the network.
convolutional network with the Euclidean loss function generally employs low learning rate because it 
easily fails to converge. As low learning rate needs much time for convergence, we employ 
an adaptive momentum approach~\cite{kingma2014adam}.

\section{Experiments}
For validating the effectiveness of the proposed architecture
to image-to-image problems, we select some of typical such
problems as skin detection, semantic segmentation, and compression
artifacts reduction.

\subsection{Experiment Setup}
Our network is implemented with the Caffe tool box~\cite{jia2014caffe}, and
GTX 980 with 4GB memory is employed both for training and testing.
The input images are subtracted by the average intensity to alleviate 
vanishing and exploding gradient problem in the training phase 
~\cite{lecun2012efficient}, and all the initial parameters for the 
convolution layers are set to zeros.
There are some hyper parameters for the proposed network which 
are decided empirically. We set the learning parameters as: initial 
learning rate = 0.001, weight decay = 0.0002.
The depth of our network is set to 20 convolution layers,
which is discussed in section 4.2.

\subsection{Application to Skin Detection}
We adopt ECU dataset~\cite{phung2005skin} for training and validation, because it is the largest reliable 
set for the skin detection. 
It contains various ethnic groups, illumination variation, and complex 
non-skin but skin-colored regions/objects which often occlude skin regions.  
The dataset consists of 4,000 images, half of which are exploited for 
training and other 2,000 images are used for evaluation. 
We adopt the precision recall (PR) and Receiver Operating Characteristic 
(ROC) curves in evaluating the performance. Moreover, the quality 
of binarized skin probability map is evaluated with four metrics: \textit{Accuracy}, 
\textit{Precision}, \textit{Recall}, and \textit{F-measure} in a fixed threshold 
which are determined to make the maximum F-measure of the dataset.

\begin{figure}[t]
	\centering
	\begin{subfigure}[]{0.5\linewidth}
		\centering
		\includegraphics[width=0.95\linewidth]{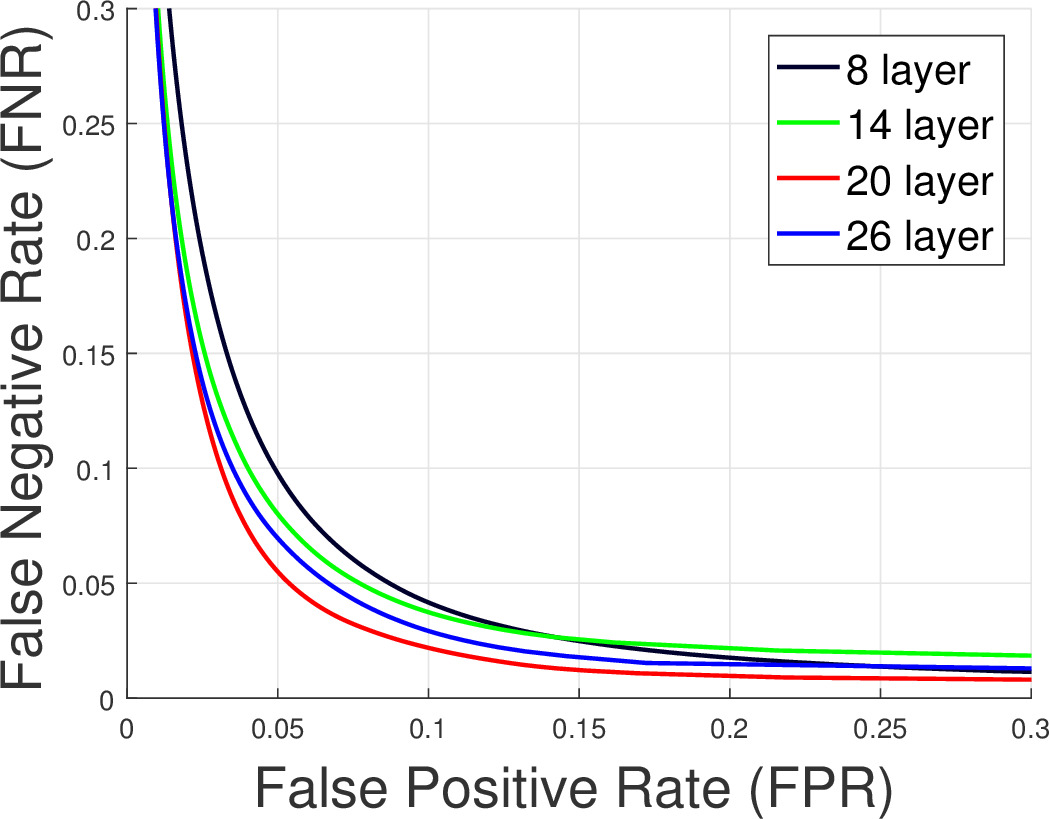}		
	\end{subfigure}%
	\begin{subfigure}[]{0.5\linewidth}
		\centering
		\includegraphics[width=0.95\linewidth]{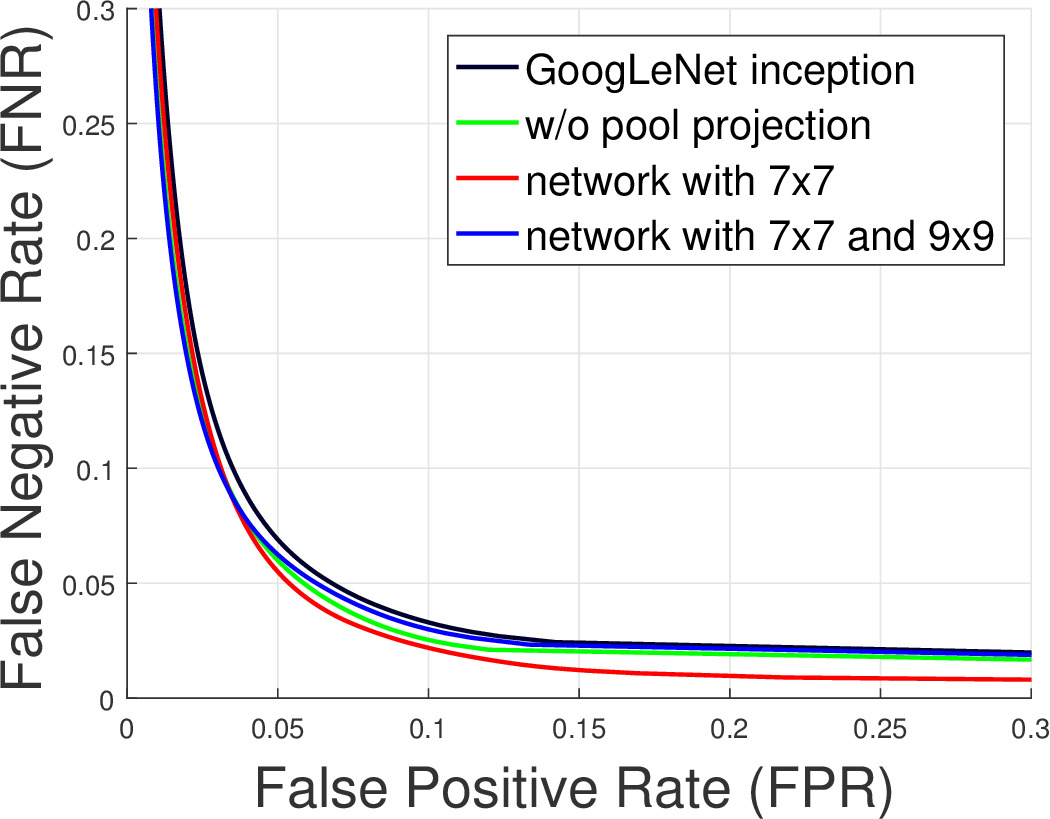}		
	\end{subfigure}
	\vspace{0.1cm}
	\caption{Quantitative comparison of proposed method with 
		variations in network structure: ROC curves depth variation (left), filter set variation (right).}	
	\label{fig:comparison_ours}		
\end{figure}    
We conduct two experiments to determine the optimal network structure 
for skin detection. First we evaluate the performance according 
to network depth: 8, 14, 20, and 26, where the number of filters and 
their sizes are all the same. We show the comparison results with ROC 
curve in Figure~\ref{fig:comparison_ours} at left the column.
It can be seen that the performance increases as the depth grows up to 20,
but the performance is lowered when 26. More fine variation in depth (not
shown here) shows that performance saturates from the depth of 20.
We think that 20 layers are enough, because this number can cover entire
region as the receptive field when the input image size is $50 \times 50$ as 
stated previously. Hence, we set the depth of layers to 20 in the rest of skin 
detection experiments. 

Second, we show the effectiveness of filter composition 
of network module (number of kernels and sizes at each layer).
We conduct experiments with four different variations.
The first module is composed of $1\times1$, $3\times3$, $5\times5$, and 
pool projection which are the same composition as GoogLeNet, 
so we denote it as \textit{GoogLeNet inception}. 
The second module is composed of $1\times1$, $3\times3$, $5\times5$ 
which removed pool projection layer in \textit{GoogLeNet inception}, 
so we denote it as \textit{w/o pool projection} and the third consists of 
$1\times1$, $3\times3$, $5\times5$, and $7\times7$ in addition, 
so we denote it as \textit{network with $7\times7$}. 
The final filter set consists of $1\times1$, $3\times3$, $5\times5$, 
$7\times7$, and $9\times9$ filters, so it is denoted as 
\textit{network with $7\times7$ and $9\times9$}. 
The comparison results with ROC curve is shown in 
Figure~\ref{fig:comparison_ours} at the right column.
It can be seen that the \textit{GoogLeNet inception} case has lower 
performance than the others. It means that pool projection does not help 
to make effective feature maps, because
some of spatial information can be lost due to max pooling.   
Since it is observed that \textit{network with $7\times7$} and 
\textit{network with $9\times9$} have similar performance on ROC 
curves, we determine to use \textit{network with $7\times7$} 
in the rest of experiments. 

\begin{figure}[h]	
	\centering
	\includegraphics[width=0.95\linewidth]{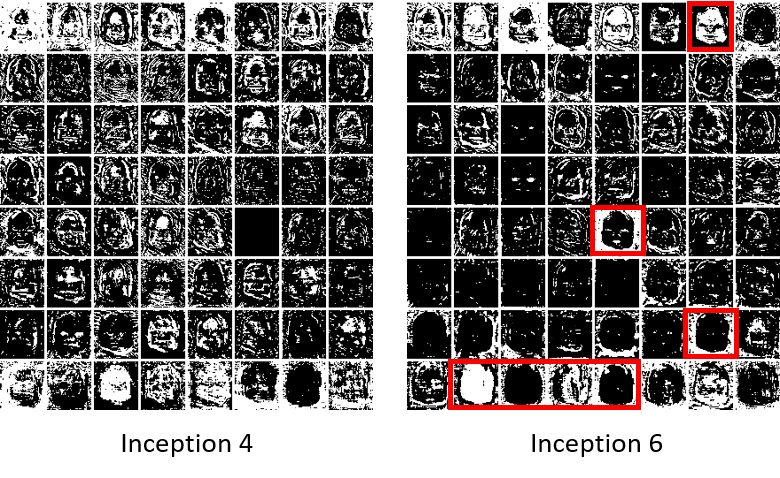}\hspace{0.5em}%
	\caption{Visualization of feature maps.}	
	\label{fig:featuremap}			
\end{figure}

We also visualize feature maps in Figure~\ref{fig:featuremap} to show how 
our inception module works. The first row shows $1\times1$ feature maps, second to fifth 
show $3\times3$, sixth to seventh $5\times5$, and eighth shows $7\times7$ feature maps. 
We can see in ``Inception 4'' feature maps that small filters usually make features 
from texture information, while large filters ($7\times7$ filter) group skin and 
non skin region from shape information.
At ``Inception 6'' feature maps, we can find that small filters can suppress small 
non skin region such as pupils and mouth, and larger filters can
find overall face as skin regions which were missed in small filters (red box).
From these observation, adding a $7\times7$ filter helps to group 
skin region more effectively.
 
\subsection{Comparison with Other Skin Detection Methods}

We compare the proposed method using benchmark skin dataset with 
other methods: Bayesian~\cite{jones2002statistical}, 
FPSD~\cite{kawulok2013fast}, DSPF~\cite{kawulok2014spatial}, 
FSD~\cite{tan2012fusion}, and LASD~\cite{hwang2013luminance}.
FPSD and DSPF are based on seed propagation over the graph
representation of images. 
FSD is the mix of dynamic threshold and Gaussian model method. 
LASD is a luminance adapted color space method which optimizes least square error. 
We also test the proposed method only with gray channel input 
to show that our method effectively exploits
structure information, and this structure is denoted as \textquotedblleft Proposed (gray)\textquotedblright.  

\begin{figure}[h]	
	\centering
	\includegraphics[width=0.48\linewidth]{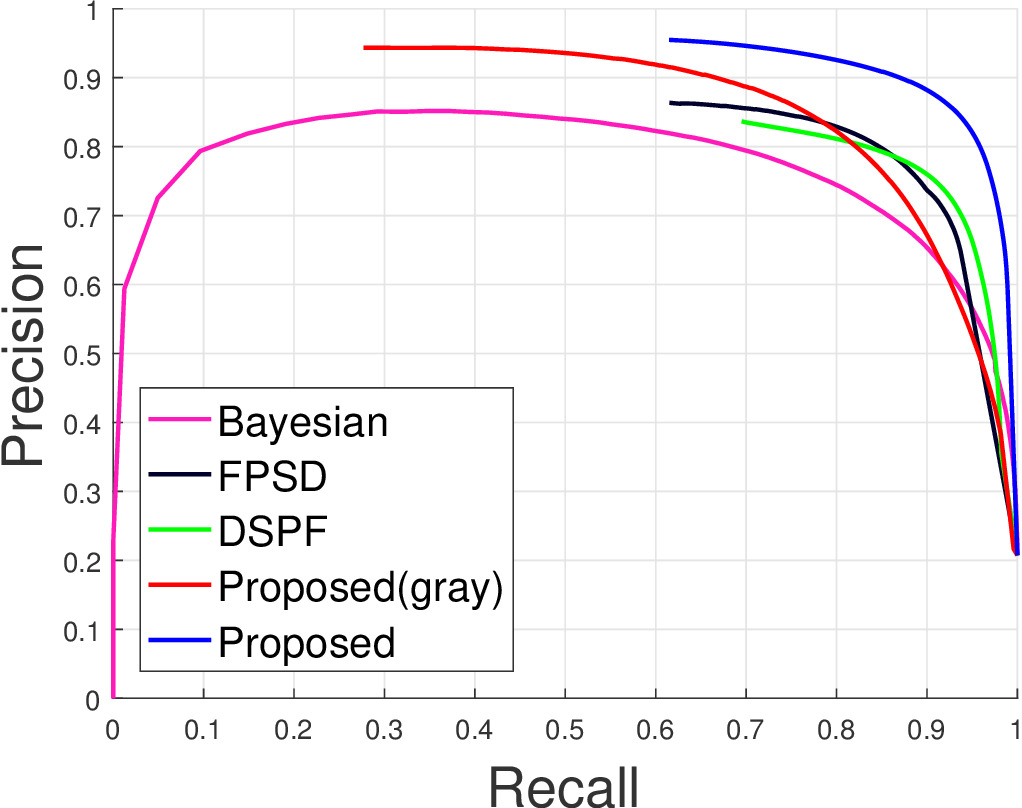}\hspace{0.5em}%
	\includegraphics[width=0.48\linewidth]{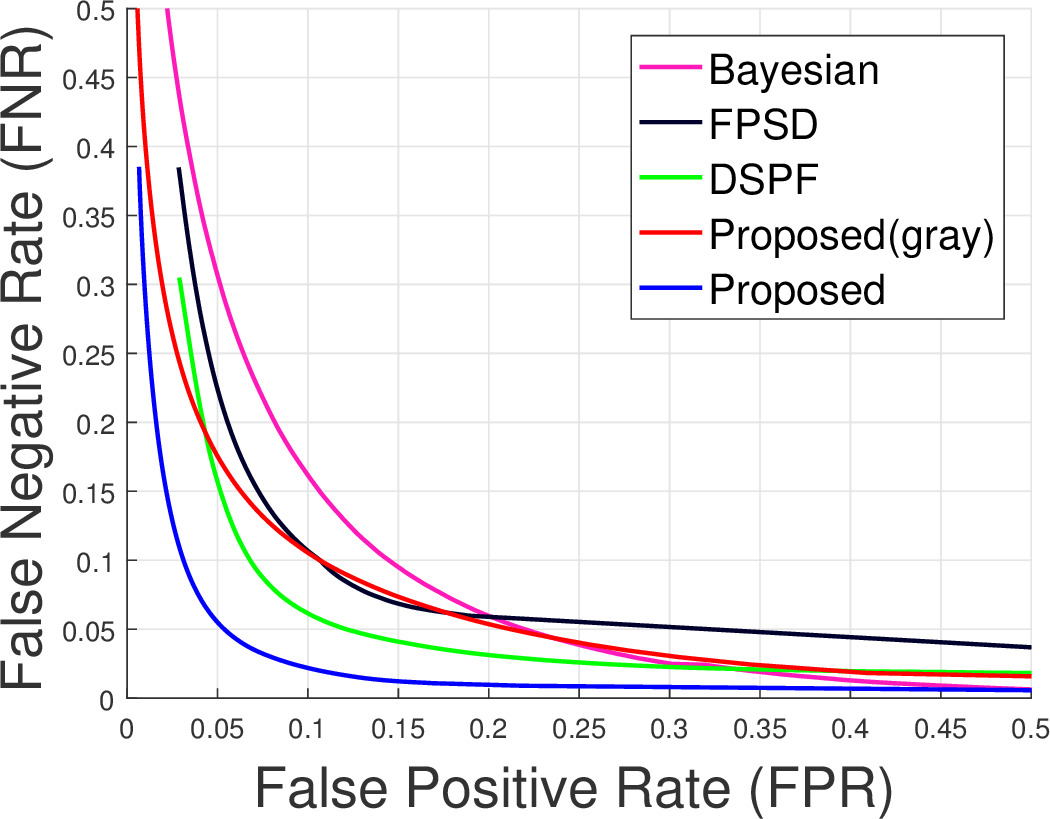}%
	\caption{Comparison of PR (left) and ROC (right) curves on ECU dataset.}	
	\label{fig:comparison_others_ECU}			
\end{figure}

\begin{table}[h]			
	\caption{Evaluation of skin detection on ECU dataset at peak F-measure.}	

	\begin{center}
		\resizebox{0.45\textwidth}{!}
		{
			\begin{tabular}{|l|c|c|c|c|c|}
				\hline
				Method &  Accuracy & Precision & Recall & F-measure\\
				\hline\hline
				Bayesian~\cite{jones2002statistical} 	&	0.8910 & 0.7292		& 0.8220	& 0.7728\\
				FPSD~\cite{kawulok2013fast}     & 0.9106		& 0.7948	& 0.8534	& 0.8231\\
				DSPF~\cite{kawulok2014spatial}   &  0.9190 		& 0.7713	& 0.8864 &  0.8249\\  								
				Proposed (gray) & 0.9276			& 0.8137	& 0.8087	& 0.8112 \\
				Proposed & \textbf{0.9562}	& \textbf{0.8720}	& \textbf{0.9122}	& \textbf{0.8917}\\
				\hline						
			\end{tabular}
		}
	\end{center}	
	\label{table:comparison_ECU}
\end{table}
We conduct test for two datasets: ECU~\cite{phung2005skin}
(without trained images) and Pratheepan~\cite{yogarajah2010dynamic}. 
The Pratheepan is 32 facial images, each of which has a single face.
First we compare the proposed method and other methods for ECU 
dataset which are plotted in Figure~\ref{fig:comparison_others_ECU}.
 Our method outperforms other methods by a large margin in terms of 
 PR and ROC curves. We can also find that using a single gray channel 
 image is also slightly better than Bayesian method which employ color 
 information. It shows that our approach uses shape of human parts (body,
 hands, and faces) as well as color information. 
 
We also evaluate the quality of binary map obtained by thresholding the above
probability map. The threshold is set as to maximize F-measure which is differently 
selected at each method. Table~\ref{table:comparison_ECU} shows this on ECU dataset, 
and our method has the best performance. 

\begin{figure}[h]
	\centering	
	\includegraphics[width=0.12\linewidth]{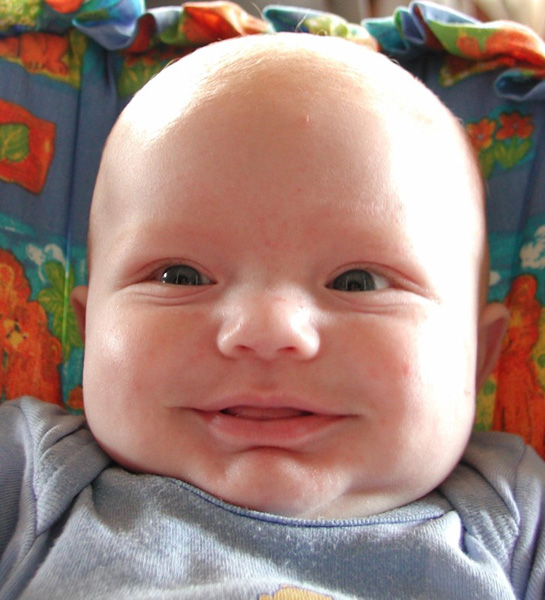}\hspace{0.1em}%
	\includegraphics[width=0.12\linewidth]{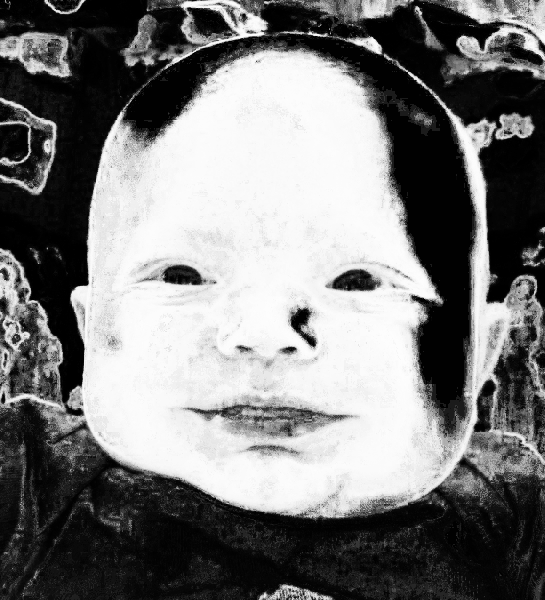}\hspace{0.1em}%
	\includegraphics[width=0.12\linewidth]{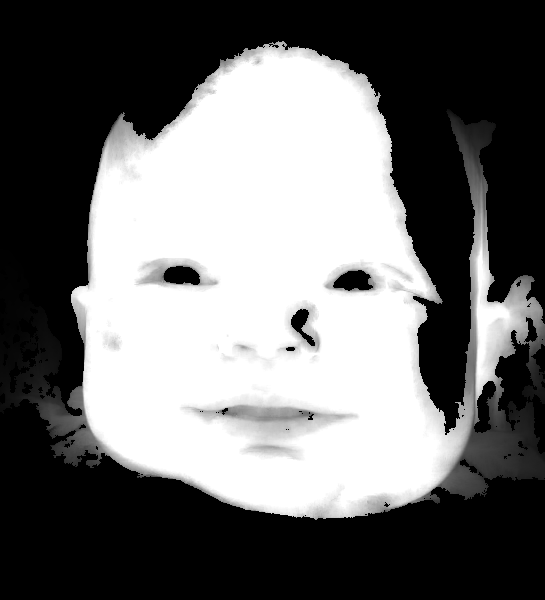}\hspace{0.1em}%
	\includegraphics[width=0.12\linewidth]{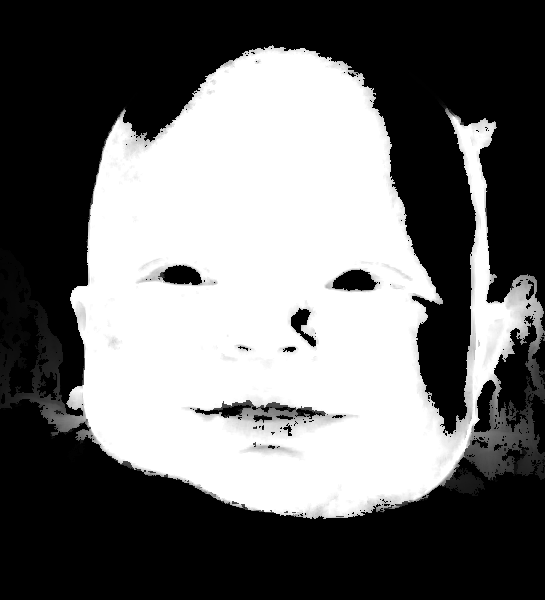}\hspace{0.1em}%
	\includegraphics[width=0.12\linewidth]{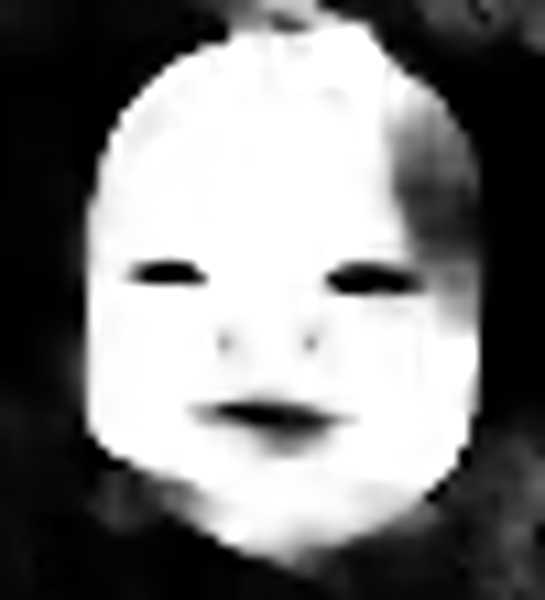}\hspace{0.1em}%
	\includegraphics[width=0.12\linewidth]{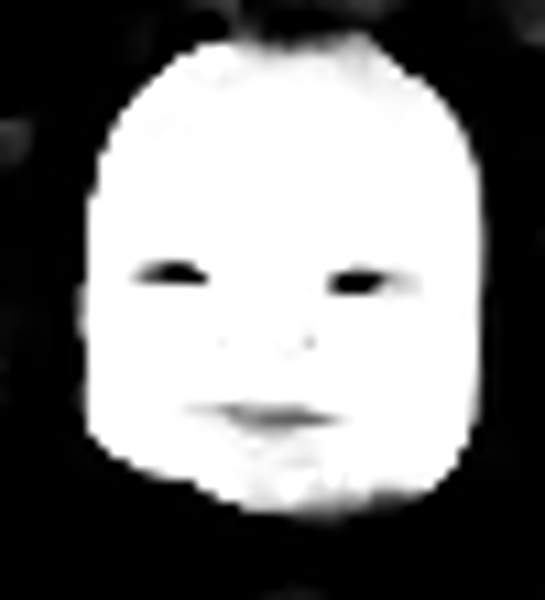}\hspace{0.1em}%
	\includegraphics[width=0.12\linewidth]{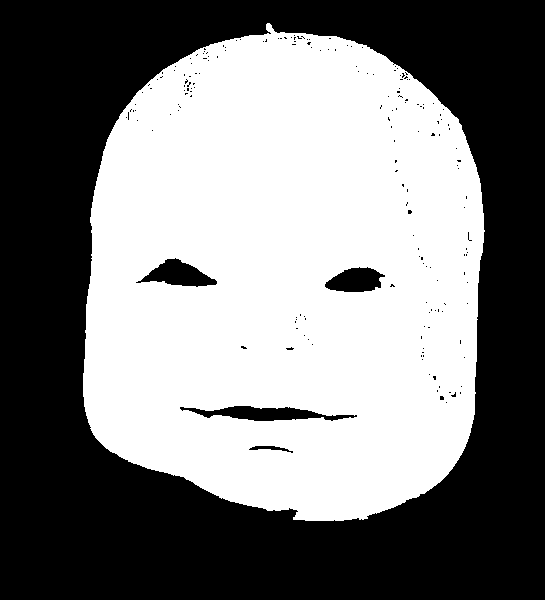}\vspace{0.1em}\\

	\includegraphics[width=0.12\linewidth]{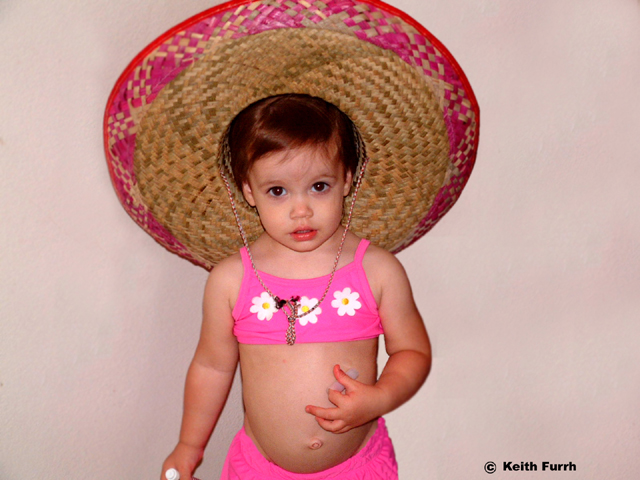}\hspace{0.1em}%
	\includegraphics[width=0.12\linewidth]{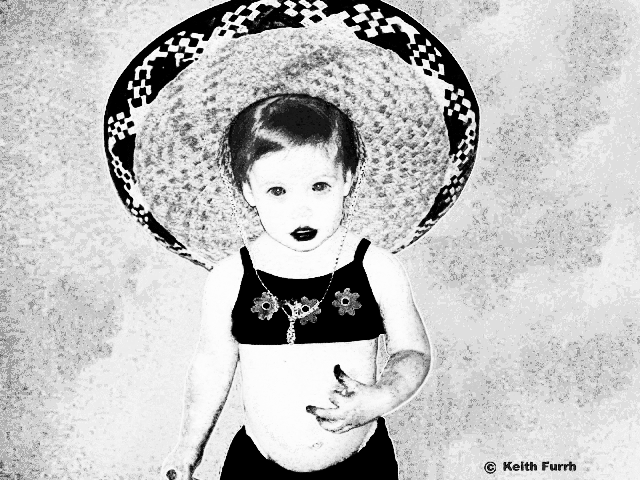}\hspace{0.1em}%
	\includegraphics[width=0.12\linewidth]{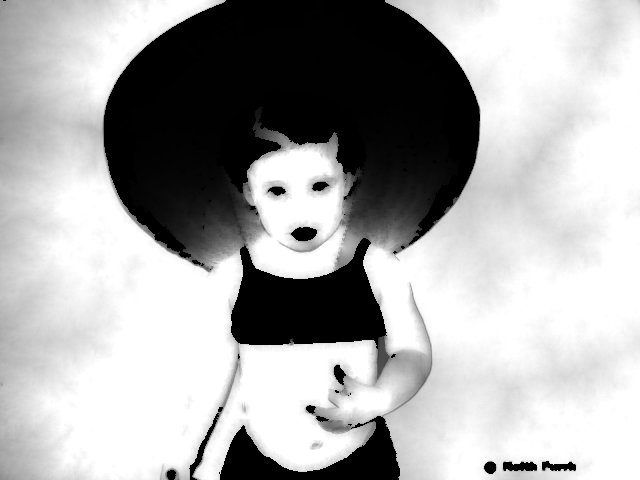}\hspace{0.1em}%
	\includegraphics[width=0.12\linewidth]{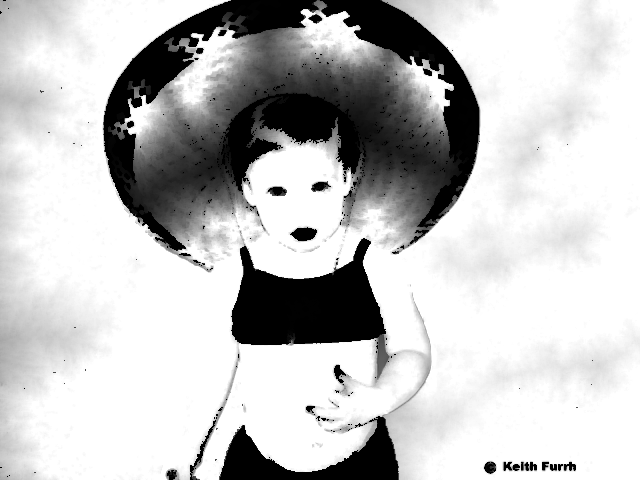}\hspace{0.1em}%
	\includegraphics[width=0.12\linewidth]{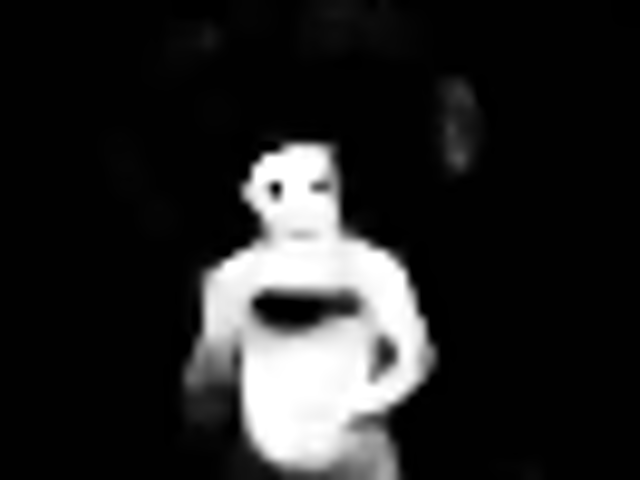}\hspace{0.1em}%
	\includegraphics[width=0.12\linewidth]{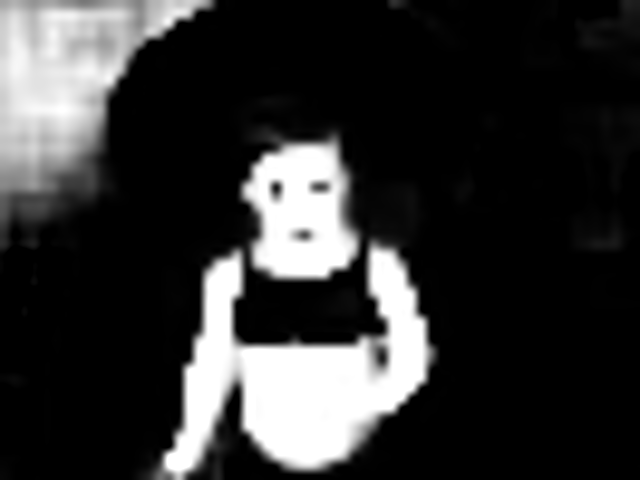}\hspace{0.1em}%
	\includegraphics[width=0.12\linewidth]{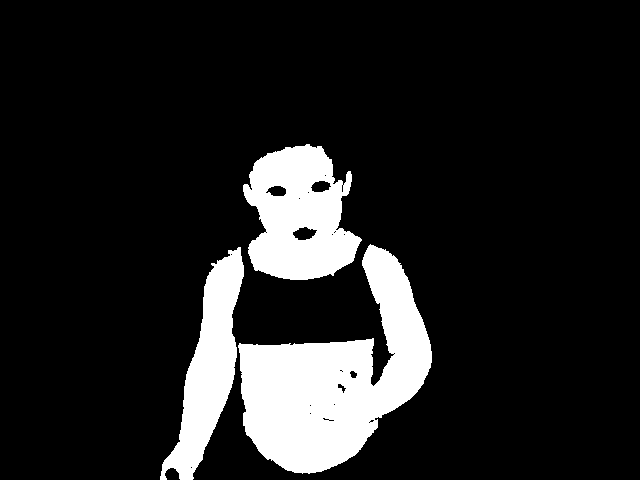}\vspace{0.1em}\\	
	
	\includegraphics[width=0.12\linewidth]{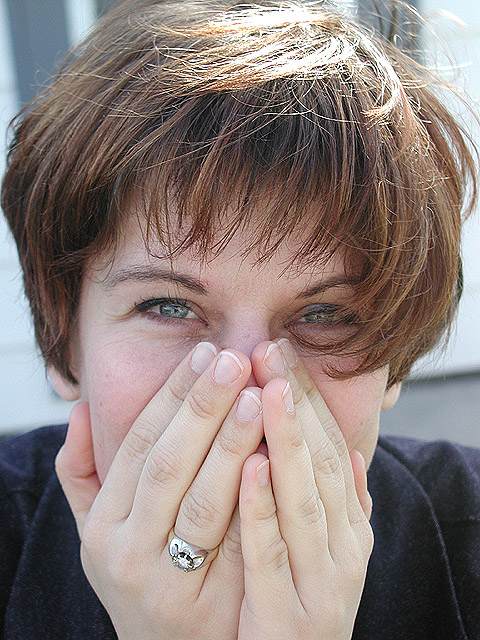}\hspace{0.1em}%
	\includegraphics[width=0.12\linewidth]{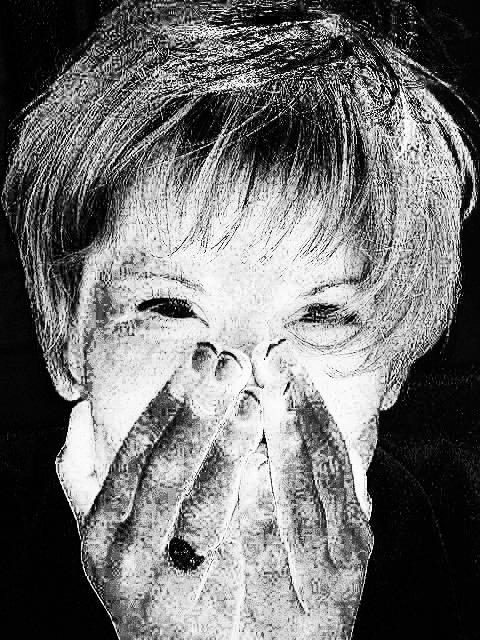}\hspace{0.1em}%
	\includegraphics[width=0.12\linewidth]{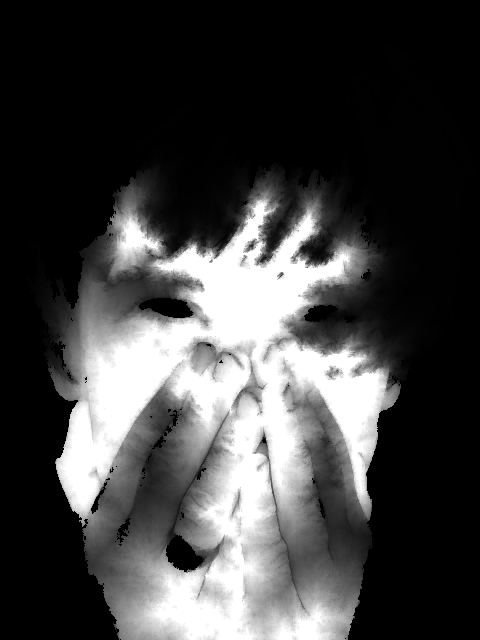}\hspace{0.1em}%
	\includegraphics[width=0.12\linewidth]{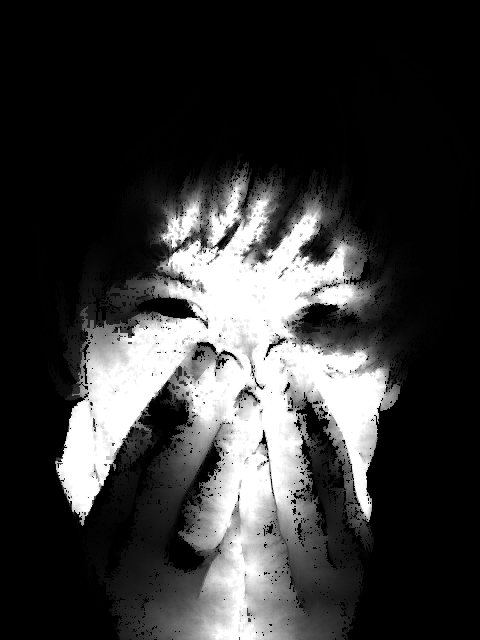}\hspace{0.1em}%
	\includegraphics[width=0.12\linewidth]{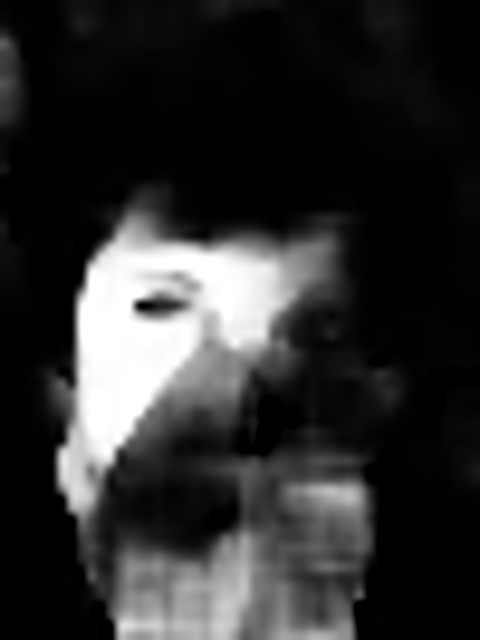}\hspace{0.1em}%
	\includegraphics[width=0.12\linewidth]{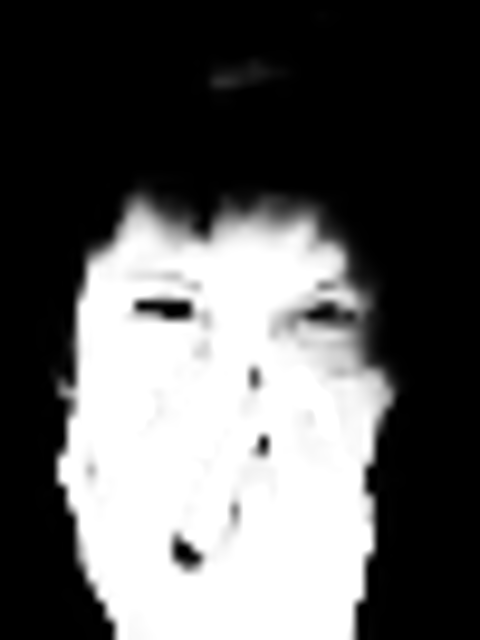}\hspace{0.1em}%
	\includegraphics[width=0.12\linewidth]{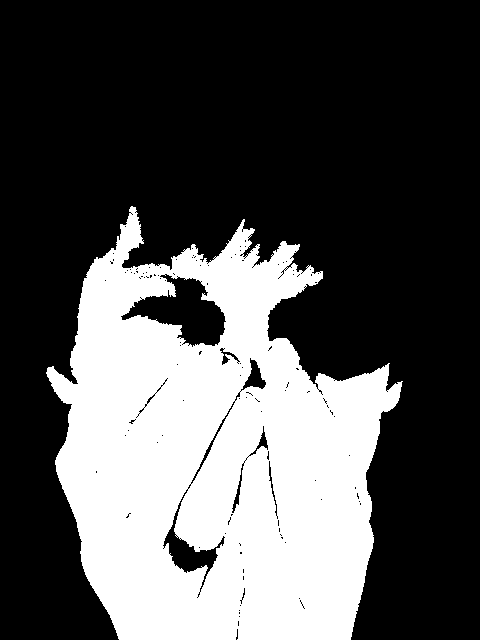}\vspace{0.1em}\\
	
\includegraphics[width=0.12\linewidth]{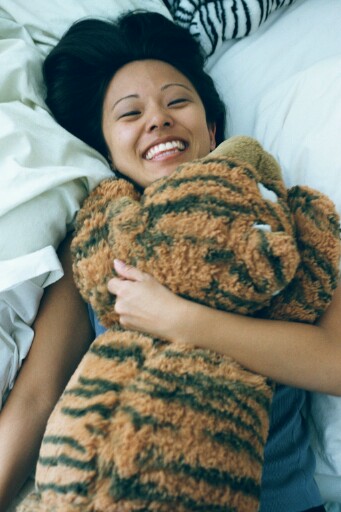}\hspace{0.1em}%
	\includegraphics[width=0.12\linewidth]{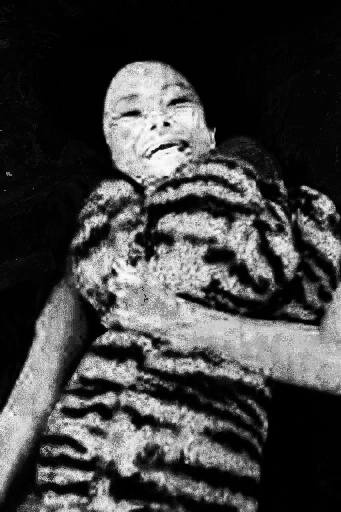}\hspace{0.1em}%
	\includegraphics[width=0.12\linewidth]{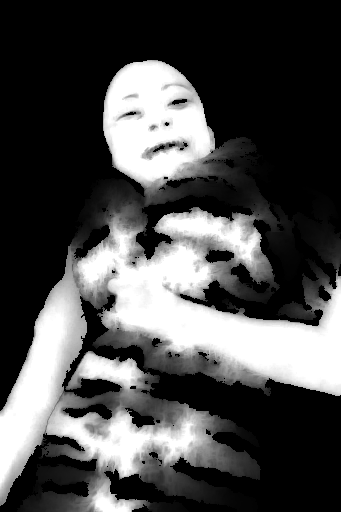}\hspace{0.1em}%
	\includegraphics[width=0.12\linewidth]{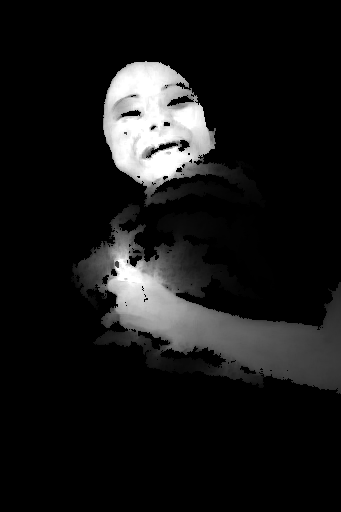}\hspace{0.1em}%
	\includegraphics[width=0.12\linewidth]{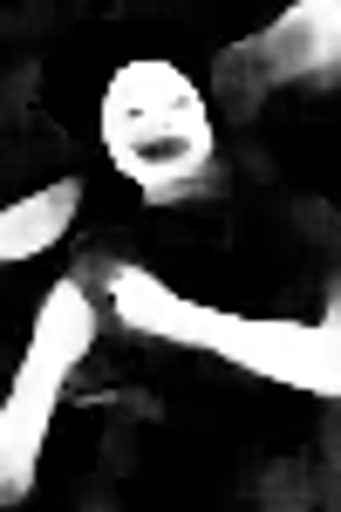}\hspace{0.1em}%
	\includegraphics[width=0.12\linewidth]{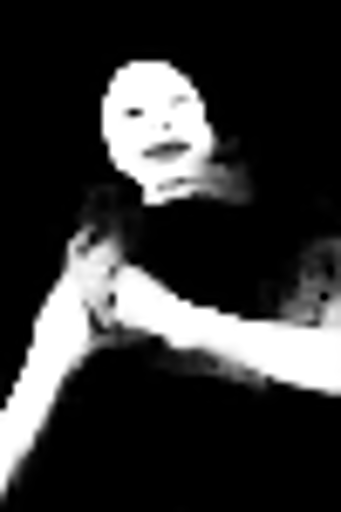}\hspace{0.1em}%
	\includegraphics[width=0.12\linewidth]{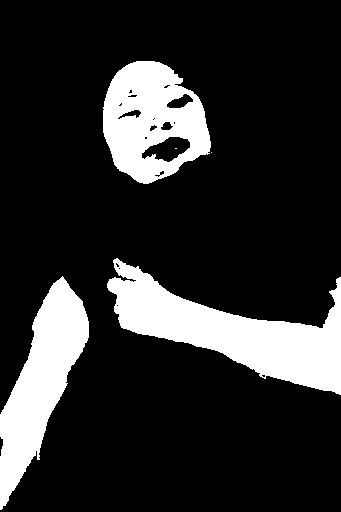}\vspace{0.1em}\\
	
	\includegraphics[width=0.12\linewidth]{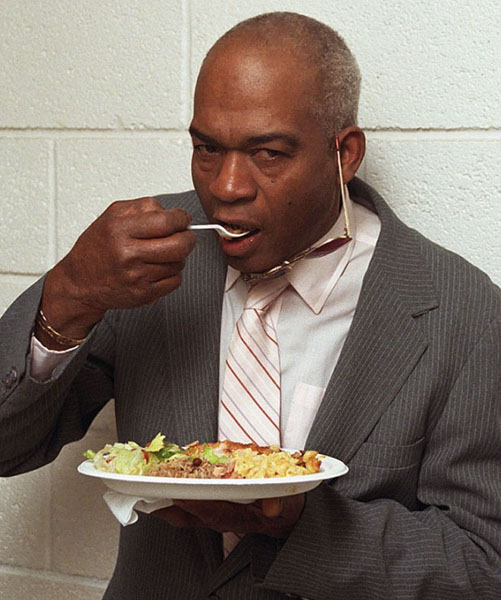}\hspace{0.1em}%
	\includegraphics[width=0.12\linewidth]{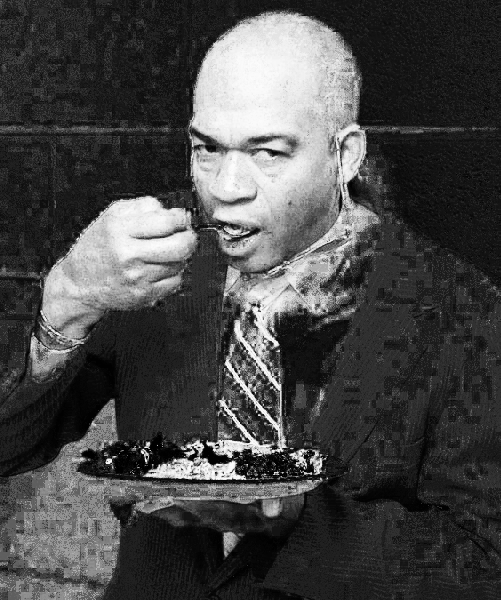}\hspace{0.1em}%
	\includegraphics[width=0.12\linewidth]{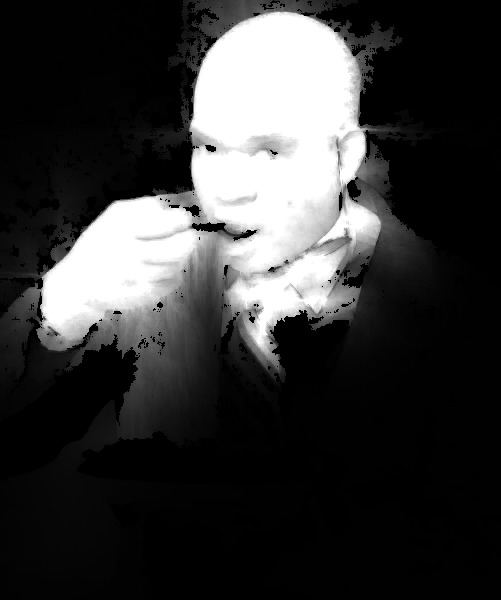}\hspace{0.1em}%
	\includegraphics[width=0.12\linewidth]{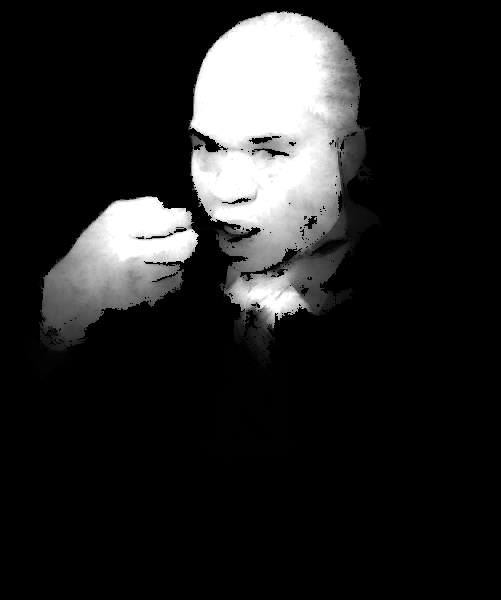}\hspace{0.1em}%
	\includegraphics[width=0.12\linewidth]{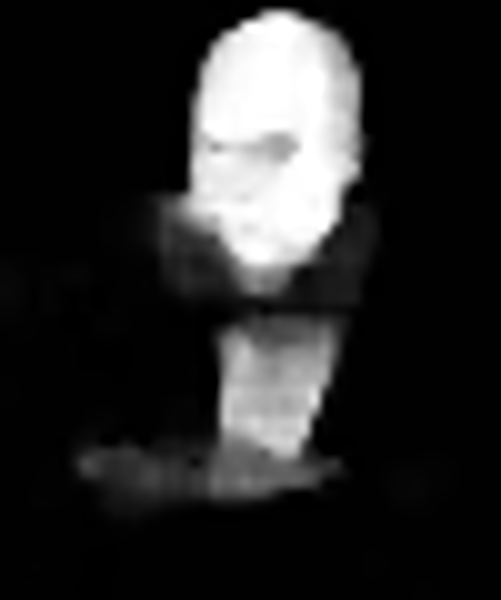}\hspace{0.1em}%
	\includegraphics[width=0.12\linewidth]{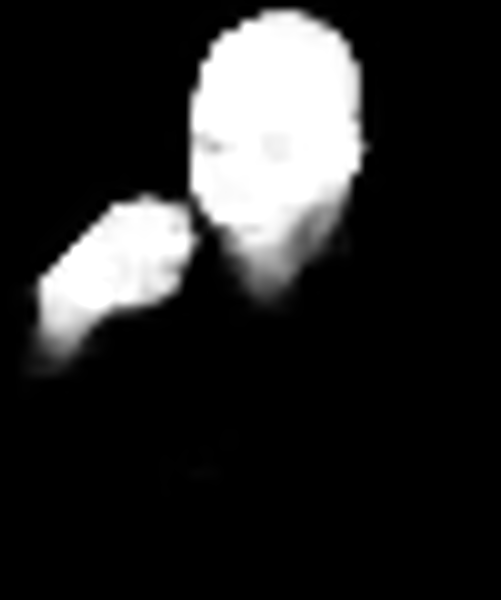}\hspace{0.1em}%
	\includegraphics[width=0.12\linewidth]{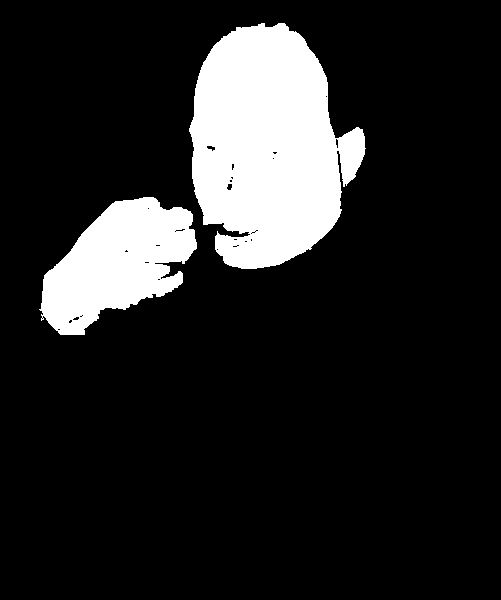}\vspace{0.1em}\\

	\caption{Subjective comparison of skin detection methods 
	on ECU dataset: (from left to right) input, Bayesian, DSPF, FPSD, proposed (gray), proposed, 
	and ground truth.}
	\label{fig:comparison_others_vis1}	
	\vspace{0.5cm}
	\includegraphics[width=0.10\linewidth]{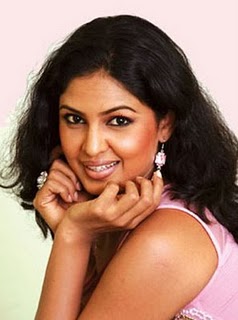}\hspace{0.1em}%
	\includegraphics[width=0.10\linewidth]{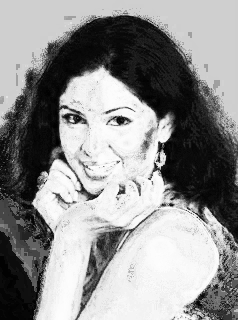}\hspace{0.1em}%
	\includegraphics[width=0.10\linewidth]{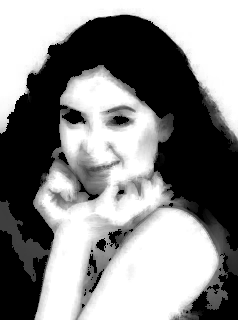}\hspace{0.1em}%
	\includegraphics[width=0.10\linewidth]{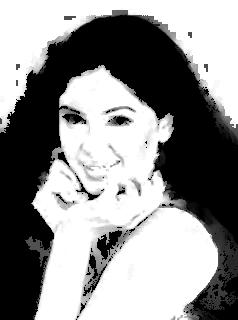}\hspace{0.1em}%
	\includegraphics[width=0.10\linewidth]{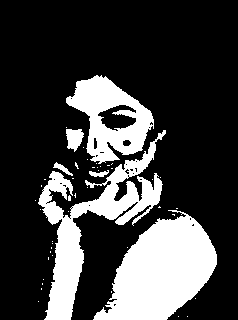}\hspace{0.1em}%
	\includegraphics[width=0.10\linewidth]{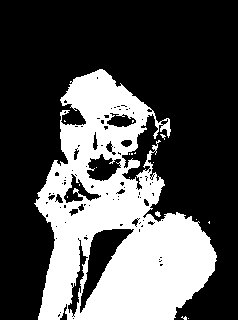}\hspace{0.1em}%
	\includegraphics[width=0.10\linewidth]{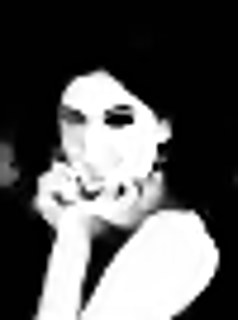}\hspace{0.1em}%
	\includegraphics[width=0.10\linewidth]{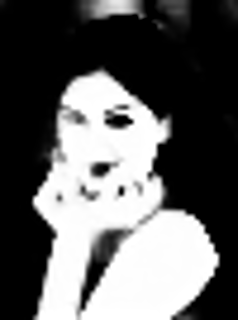}\hspace{0.1em}%
	\includegraphics[width=0.10\linewidth]{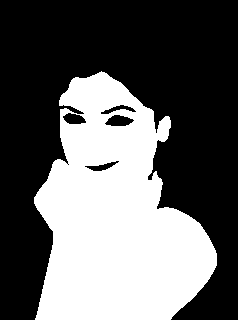}\hspace{0.1em}%
	
	\includegraphics[width=0.10\linewidth]{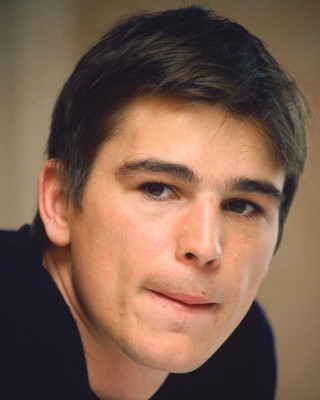}\hspace{0.1em}%
	\includegraphics[width=0.10\linewidth]{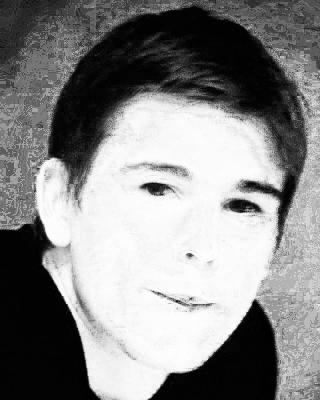}\hspace{0.1em}%
	\includegraphics[width=0.10\linewidth]{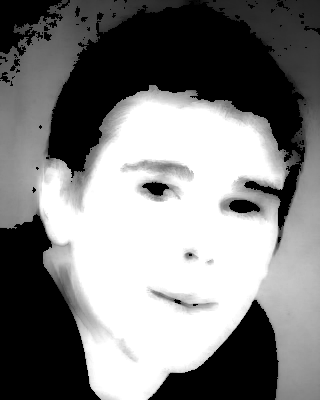}\hspace{0.1em}%
	\includegraphics[width=0.10\linewidth]{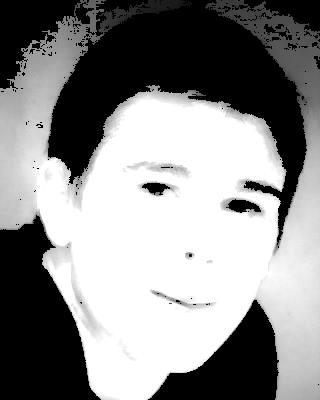}\hspace{0.1em}%
	\includegraphics[width=0.10\linewidth]{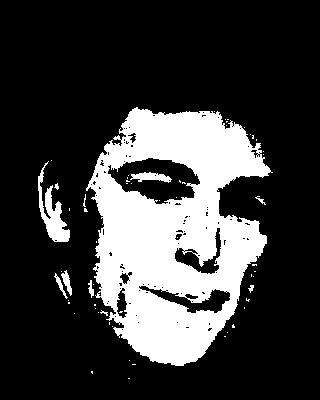}\hspace{0.1em}%
	\includegraphics[width=0.10\linewidth]{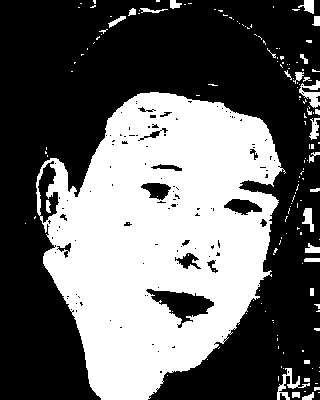}\hspace{0.1em}%
	\includegraphics[width=0.10\linewidth]{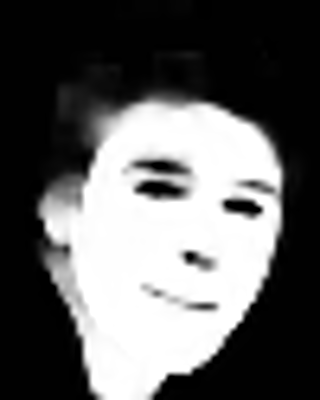}\hspace{0.1em}%
	\includegraphics[width=0.10\linewidth]{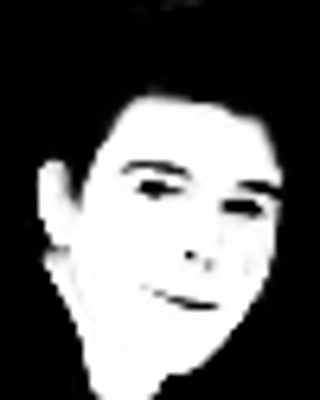}\hspace{0.1em}%
	\includegraphics[width=0.10\linewidth]{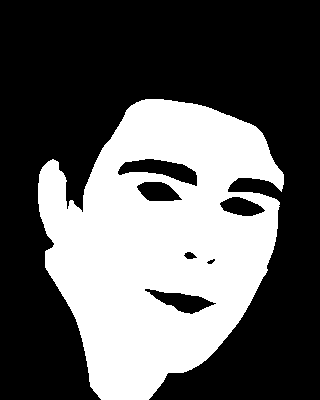}\hspace{0.1em}%

\caption{Subjective comparison of other methods on Prateepan dataset: (from left to right) input, Bayesian, DSPF, FPSD, FSD, LASD, proposed (gray), proposed, and ground truth.}
\label{fig:comparison_others_vis2}	
\end{figure}

Subjective comparison is presented in Figure~\ref{fig:comparison_others_vis1}.
It shows that our method robustly detects the skins of various races, 
rejects skin-like background region (non-skin region), and works robustly 
to illumination variations. Specifically, at the first row of Figure~\ref{fig:comparison_others_vis1},
the illumination variation on the baby's face is very severe, which lowered the
performance of other methods, while our method works robustly as it uses color and
shape information effectively as illustrated in Figure~\ref{fig:featuremap}.
It is also shown that other methods often fail to suppress skin-like background 
clutter (row 2,3, and 4), whereas our method can overcome these difficulties very well.

\begin{figure}[h]	
	\centering
	\includegraphics[width=0.48\linewidth]{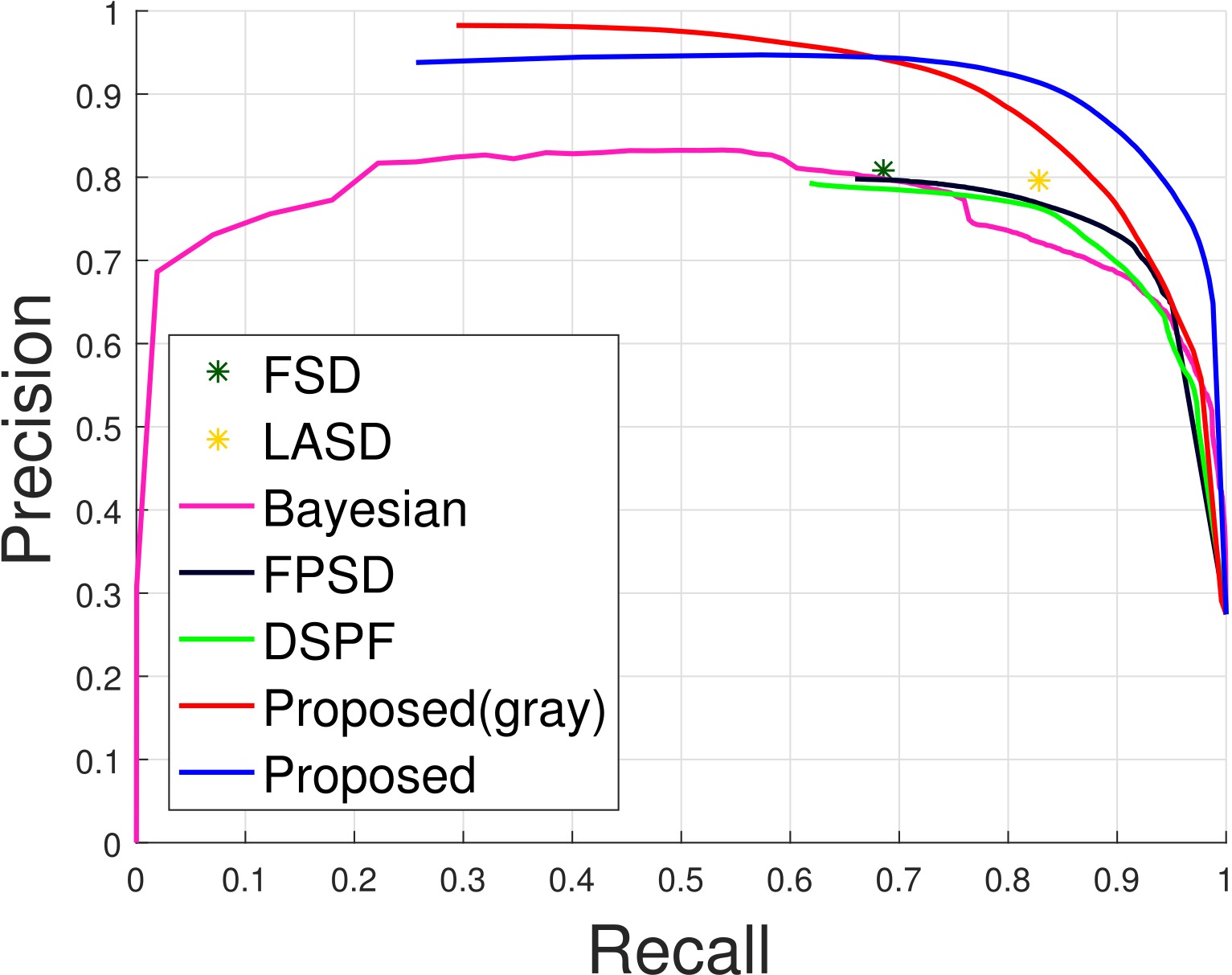}\hspace{0.5em}%
	\includegraphics[width=0.48\linewidth]{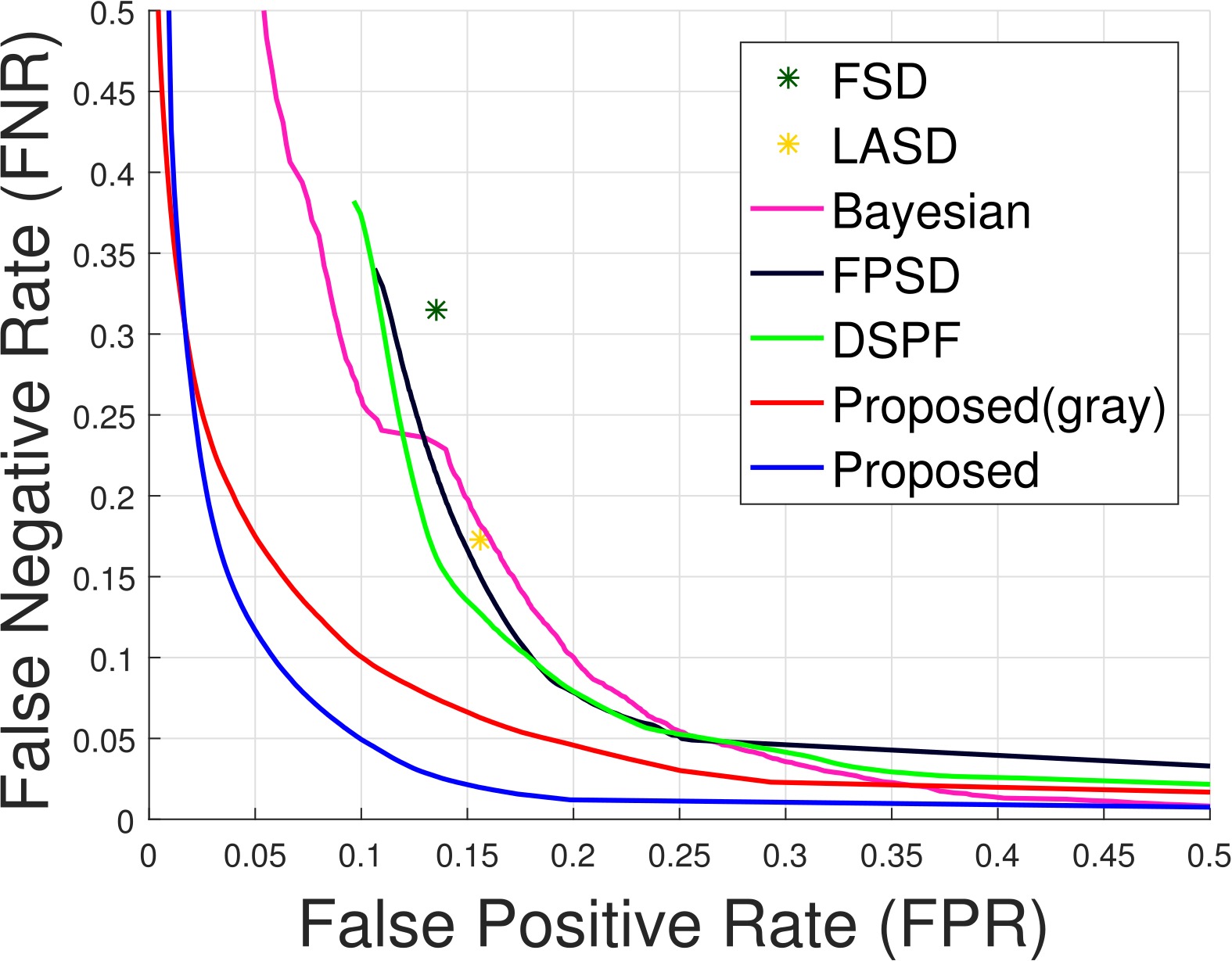}%
	\caption{Comparison of PR (left) and ROC (right) curves on Pratheepan dataset.}	
	\label{fig:comparison_others_PRA}			
\end{figure}

\begin{table}[h]	
	\caption{Evaluation of skin detection on Prateepan dataset at peak F-measure.}
	
	\begin{center}		
		\resizebox{0.45\textwidth}{!}
		{
			\begin{tabular}{|l|c|c|c|c|c|}
				\hline
				Method &  Accuracy & Precision & Recall & F-measure\\
				\hline\hline
				Bayesian~\cite{jones2002statistical}   &	0.8237  & 0.6881		& 0.8972	& 0.7788 	\\
				FSD~\cite{tan2012fusion} &	0.8255  & 0.8077		& 0.6851	& 0.7414\\
				LASD~\cite{hwang2013luminance} &	 0.8361 & 0.7954& 0.8275 & 0.8111\\
				FPSD~\cite{kawulok2013fast}     &  0.8419		& 0.7837	& \textbf{0.8991}	& 0.8070	\\
				DSPF~\cite{kawulok2014spatial}    &   0.8521  		& 0.7543	& 0.8436	& 0.7964	\\								
				Proposed (gray)  & 0.9211	& 0.8296 & 0.8379 & 0.8337 \\
				Proposed & \textbf{0.9483}	& \textbf{0.9003} & 0.8912 & \textbf{0.8957} \\
				\hline						
			\end{tabular}
		}
	\end{center}	
	\label{table:comparison_Prateepan}
\end{table}

Comparison on Pratheepan dataset is shown in Figure~\ref{fig:comparison_others_PRA},
where it can be seen that the proposed method has the best performance on both 
PR and ROC curves. Moreover, it has much better performance than others even 
in the case of using only gray channel image.
Table~\ref{table:comparison_Prateepan} also shows that proposed method yields
good results in terms of many performance measures. 
Finally, the subjective comparison for Pratheepan dataset is
shown in Figure~\ref{fig:comparison_others_vis2}.

\subsection{Application to Semantic Segmentation}

We also apply the proposed network to semantic segmentation problem, where
the only difference from the above skin detection network is the loss function.
Specifically, we use softmax as a loss function and the other parameters
(depth, number of filters, learning rate, etc) are kept the same as the skin detection work.
We train and evaluate proposed architecture using CamVid~\cite{brostow2009semantic} 
set which is a 11 class set (mainly road, building, sky, and car).
We train our network employing 367 training images and test with 233 RGB images 
at 360 $\times$ 480 resolution.
It is compared with FCN~\cite{long2015fully} and SegNet~\cite{badrinarayanan2015segnet} 
which are the state of the art semantic segmentation convolutional networks.
We adopt three commonly employed evaluation measure: \textit{Accuracy} for global 
pixel-wise accuracy, \textit{Class Mean} for class average accuracy 
, and \textit{I/U} is the mean of intersection over union.

\begin{figure*}[t]	
	\centering
	\includegraphics[width=0.19\linewidth]{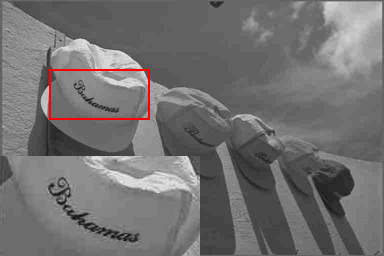}\hspace{0.1em}%
	\includegraphics[width=0.19\linewidth]{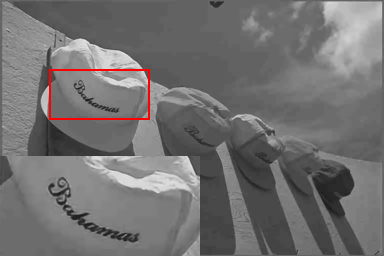}\hspace{0.1em}%
	\includegraphics[width=0.19\linewidth]{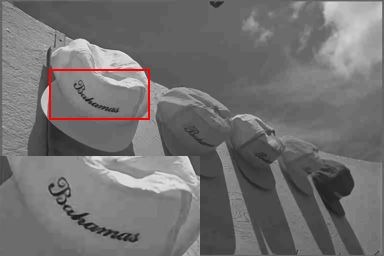}\hspace{0.1em}%
	\includegraphics[width=0.19\linewidth]{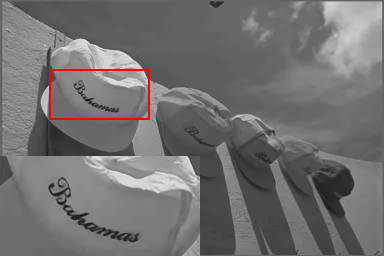}\hspace{0.1em}%
	\includegraphics[width=0.19\linewidth]{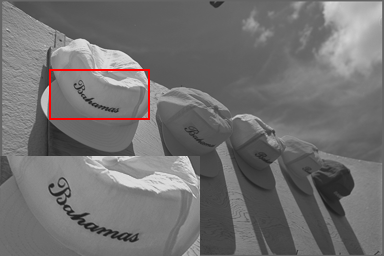}\hspace{0.1em}\vspace{0.1em}\\	
	
	\includegraphics[width=0.19\linewidth]{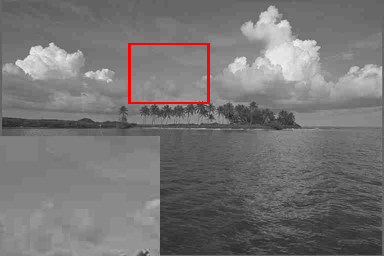}\hspace{0.1em}%
	\includegraphics[width=0.19\linewidth]{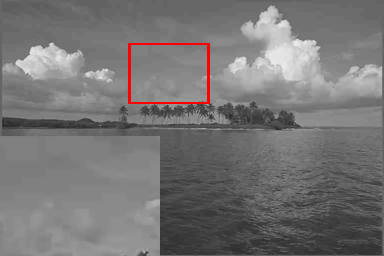}\hspace{0.1em}%
	\includegraphics[width=0.19\linewidth]{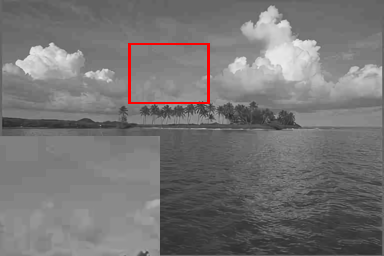}\hspace{0.1em}%
	\includegraphics[width=0.19\linewidth]{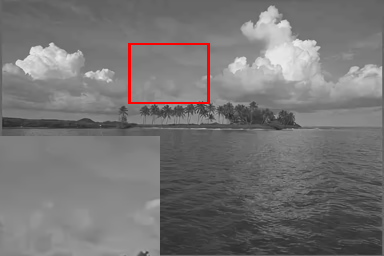}\hspace{0.1em}%
	\includegraphics[width=0.19\linewidth]{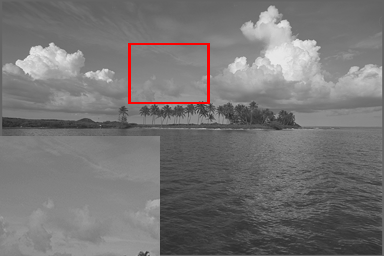}\hspace{0.1em}\vspace{0.1em}\\	
	
	\caption{Subjective comparison of the proposed method with others: 
		(from left to right) JPEG, AR-CNN, RAR-CNN, proposed, 
		and original.}
	\label{fig:comparison_arcnn}	
\end{figure*}

\begin{figure}[h]
	\centering	
	\includegraphics[width=0.3\linewidth]{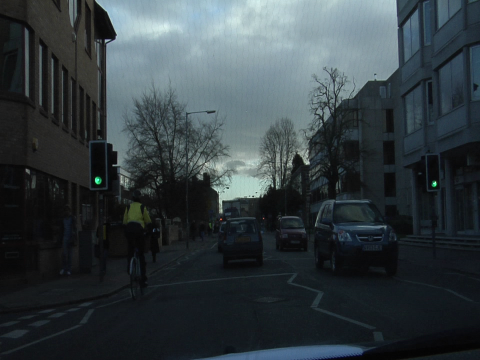}\hspace{0.3em}%
	\includegraphics[width=0.3\linewidth]{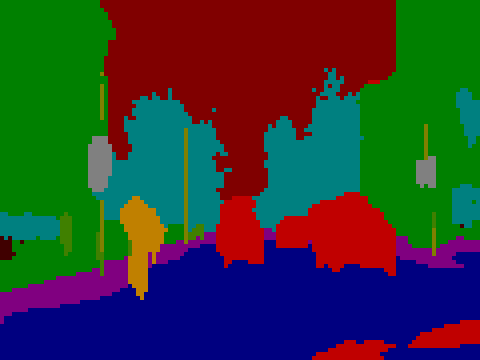}\hspace{0.3em}%
	\includegraphics[width=0.3\linewidth]{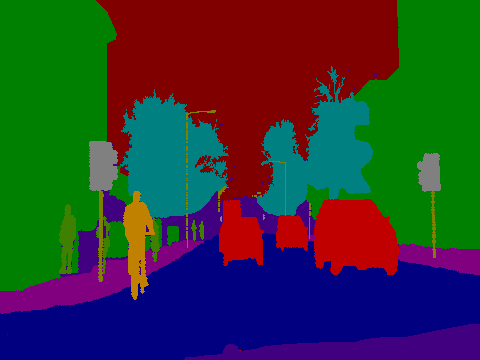}\vspace{0.1em}\\	
	
	\includegraphics[width=0.3\linewidth]{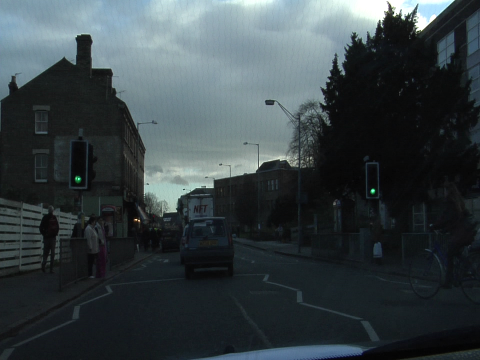}\hspace{0.3em}%
	\includegraphics[width=0.3\linewidth]{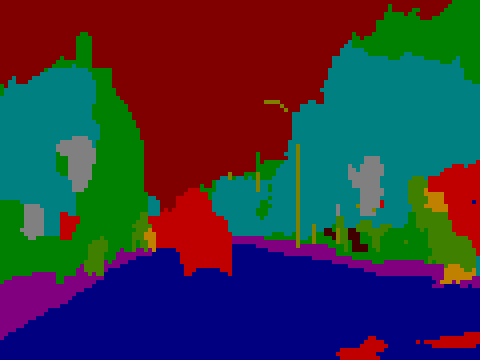}\hspace{0.3em}%
	\includegraphics[width=0.3\linewidth]{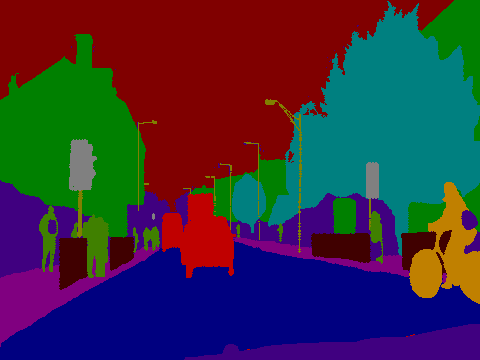}\vspace{0.1em}\\	
	
	\includegraphics[width=0.3\linewidth]{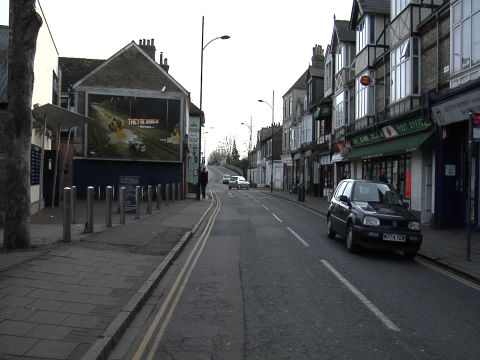}\hspace{0.3em}%
	\includegraphics[width=0.3\linewidth]{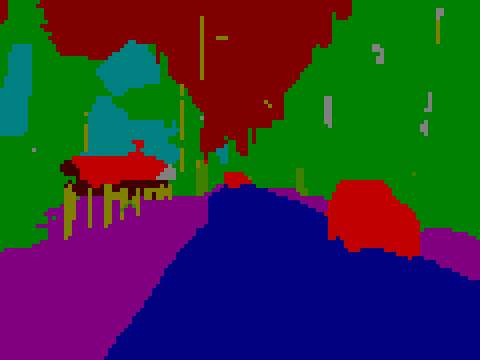}\hspace{0.3em}%
	\includegraphics[width=0.3\linewidth]{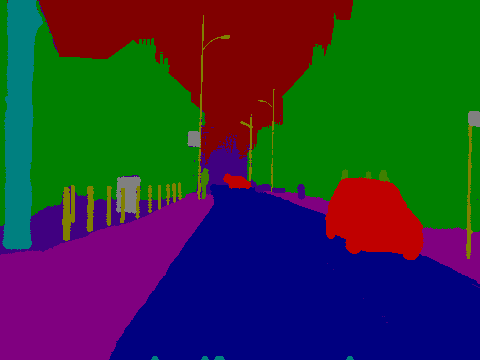}\vspace{0.1em}\\	
	
	\caption{Visualization on CamVid dataset: (from left to right) input, proposed, and ground truth.}
	\label{fig:visualization_cam}	
\end{figure}

\begin{table}[h]
	\caption{Evaluation of semantic segmentation on CamVid dataset.}	
	\begin{center}
		\resizebox{0.35\textwidth}{!}
		{
			\begin{tabular}{|l|c|c|c|c|c|}
				\hline
				Method &  Accuracy & Class Mean & I/U\\
				\hline\hline
				FCN~\cite{long2015fully} 	&	0.839 & 0.556		& 0.450\\
				SegNet~\cite{badrinarayanan2015segnet}     & 0.842		& 0.565	& 0.477	\\
				Proposed    &  \textbf{0.855} 		& \textbf{0.651}	& \textbf{0.517} \\				
				\hline						
			\end{tabular}
		}
	\end{center}	
	\label{table:comparison_CamVid}
\end{table}
Quantitative results in Table~\ref{table:comparison_CamVid} show that 
the proposed method 
yields better performance than the others in terms of all measure metrics.
Visualization of output is presented in Figure~\ref{fig:visualization_cam}. 
Proposed method classifies road and car very well, but it has limitations when the trees occlude buildings.

\subsection{Application to Compression Artifacts Reduction}
Our network is also applied to reducing the compression artifacts,
especially the compression noise in JPEG images.
The number of training images is 400 and test dataset is 5 classical 
images and LIVE1 which contains 29 images.
We compare the proposed algorithm to SA-DCT~\cite{foi2007pointwise} and convolutional network such as AR-CNN~\cite{dong2015compression} and RAR-CNN~\cite{svoboda2016compression}.
SA-DCT is deblocking oriented method at frequency domain which is regarded as the state-of-the-art method before using a convolutional network.
AR-CNN consist of four convolution layers and RAR-CNN is composed of 
four convolution layer which is trained with residual learning.
\begin{table}[h]	
	\caption{Image reconstruction quality on 5 classical validation images for JPEG quality 10 and 20.}
	\begin{center}
		\resizebox{0.4\textwidth}{!}
		{
			\begin{tabular}{|l|c|c|c|c|c|c|}
				\hline
				\multirow{2}{*}{Methods} &\multicolumn{2}{c}{Q = 10} & \multicolumn{2}{c}{Q = 20} \vline \\\cline{2-5} 												
				&  \textit{PSNR} & \textit{SSIM}	&	\textit{PSNR}	& \textit{SSIM}\\\hline\hline				
				JPEG  	&	27.82 & 0.780		& 30.12 & 0.854\\
				SA-DCT~\cite{foi2007pointwise}     & 28.88		& 0.807	& 30.92 & 0.8663	\\
				AR-CNN~\cite{dong2015compression}     & 29.04		& 0.811	& 31.16& 0.869	\\							
				Proposed  & \textbf{29.34} &  \textbf{0.820} 		& \textbf{31.51}	& \textbf{0.876} \\				
				\hline						
			\end{tabular}
		}
	\end{center}	
	\label{table:comparison_5classic_arcnn}
\end{table}
\begin{table}[h]	
	\caption{Image reconstruction quality on LIVE1 validation dataset for JPEG quality 10 and 20.}
	\begin{center}
		\resizebox{0.4\textwidth}{!}
		{
			\begin{tabular}{|l|c|c|c|c|c|c|}
				\hline
				\multirow{2}{*}{Methods} &\multicolumn{2}{c}{Q = 10} & \multicolumn{2}{c}{Q = 20} \vline \\\cline{2-5} 												
				&  \textit{PSNR} & \textit{SSIM}	&	\textit{PSNR}	& \textit{SSIM}\\\hline\hline				
				JPEG  	&	27.77 & 0.791		& 30.07 & 0.868\\
				SA-DCT~\cite{foi2007pointwise}     & 28.65		& 0.809	& 30.81 & 0.878	\\
				AR-CNN~\cite{dong2015compression}     & 28.98		& 0.822	& 31.29& 0.887	\\
				RAR-CNN~\cite{svoboda2016compression}     & 29.08		& 0.824	& 31.42	&0.891\\				
				Proposed  & \textbf{29.25} &  \textbf{0.828} 		& \textbf{31.60}	& \textbf{0.894} \\				
				\hline						
			\end{tabular}
		}
	\end{center}	
	\label{table:comparison_arcnn}
\end{table}
Quantitative results evaluated on 5 classical images and LIVE1 dataset are presented in Table~\ref{table:comparison_5classic_arcnn} and~\ref{table:comparison_arcnn}, which
show that the proposed method yields the best performance for the 
compressed images at the quality factor of 10 and 20 in 
terms of PSNR and SSIM.
Subjective results are shown in Figure~\ref{fig:comparison_arcnn}, where
enlarged image can be found in the supplementary file.

\section{Conclusions}
We have proposed a network-in-network structure based on the 
inception module of GoogLeNet, which can be effectively used for
the problems of reconstructing an image as the output.
Specifically, the pool projection is removed from the original inception 
module in order not to decimate the features, and $7\times7$ convolutional network
is added instead to keep the receptive field wide.
The proposed architecture is
applied to skin detection, semantic segmentation
and compression artifacts reduction, and it is shown to yield competent results compared
with the state-of-the-art methods. 
We believe that the proposed network widens the application
areas of inception-like modules to the image-to-image problems, \ie,
it can be adopted to many other image restoration problems with
some modification and tuning.

{\small
	\bibliographystyle{ieee}
	\bibliography{egbib}
}

\end{document}